\pdfoutput=1

\documentclass[11pt]{article}

\usepackage[final]{acl}

\usepackage{times}
\usepackage{latexsym}

\usepackage[T1]{fontenc}

\usepackage[utf8]{inputenc}

\usepackage{microtype}

\usepackage{inconsolata}

\usepackage{hyperref}
\usepackage{url}
\usepackage{graphicx}  
\usepackage{xspace}
\usepackage{multirow} 
\usepackage{booktabs} 
\usepackage{arydshln} 
\usepackage{siunitx}
\usepackage{subcaption} 
\usepackage{enumitem} 
\usepackage{xcolor}
\usepackage{colortbl}

\usepackage{makecell} 
\usepackage{wrapfig} 

\usepackage{pifont} 
\usepackage[most]{tcolorbox}
\definecolor{darkgreen}{RGB}{0,160,0}
\newcommand{\notcheckmark}{\textcolor{black}{\bcmark\kern-1.1ex\raisebox{.7ex}{\rotatebox[origin=c]{125}{--}}}\color{black}}
\newcommand{\bcmark}{\color{blue}{\ding{51}}}%
\newcommand{\cmark}{\color{darkgreen}{\ding{51}}}%
\newcommand{\xmark}{\color{red}{\ding{55}}}%

\usepackage{xcolor}  
\definecolor{Orange}{HTML}{e9864c}
\definecolor{Green}{HTML}{70ad47}

\newcommand{\data}{\textsc{ReachQA}\xspace}

\title{Distill Visual Chart Reasoning Ability from LLMs to MLLMs}

\author{Wei He$^1$\thanks{Equal contribution. Work done during Wei He's internship at Tencent. $^\dag$Corresponding authors.}\ , 
\ Zhiheng Xi$^{1*}$, 
\ Wanxu Zhao$^{1*}$, 
\ Xiaoran Fan$^1$, 
\ Yiwen Ding$^1$, 
\\
\textbf{Zifei Shan}$^2$, 
\ \textbf{Tao Gui}$^{1,3\dag}$, 
\ \textbf{Qi Zhang}$^{1,4\dag}$, 
\ \textbf{Xuanjing Huang}$^{1,4}$ 
\\
$^1$School of Computer Science, Fudan University\ \ \ $^2$Weixin Group, Tencent
\\
$^3$Shanghai Innovation Institute\ \ \ $^4$Shanghai Key Lab of Intelligent Information Processing
\\
\texttt{whe23@m.fudan.edu.cn, \{tgui,qz\}@fudan.edu.cn}
}

\begin{document}
\maketitle

\begin{abstract}
Solving complex chart Q\&A tasks requires advanced visual reasoning abilities in multimodal large language models (MLLMs), including recognizing key information from visual inputs and conducting reasoning over it.
While fine-tuning MLLMs for reasoning is critical, collecting and annotating charts and questions is expensive, hard to scale, and often results in low-quality annotations.
To address this, we propose \textit{Code-as-Intermediary Translation} (CIT), a cost-effective, efficient and scalable data synthesis method for distilling visual reasoning abilities \textbf{from LLMs to MLLMs}.
The code serves as an intermediary that translates visual chart representations into textual representations, enabling language models to understand cross-modal information and generate reasoning chains accordingly.
In this way, we can employ text-based synthesizing techniques to expand chart-plotting code and generate high-quality Q\&A pairs for training models. 
This produces \textbf{\data}, a dataset containing $3\text{k}$ \underline{rea}soning-intensive \underline{ch}arts and $20\text{k}$ Q\&A pairs to enhance both recognition and reasoning abilities of MLLMs.
Experiments show that models fine-tuned with \data not only perform well on chart-related tasks but also show performance gains on general reasoning benchmarks.
\end{abstract}

\section{Introduction}

\begin{figure}[!ht]
    \begin{center}
    \includegraphics[width=0.48\textwidth]
    {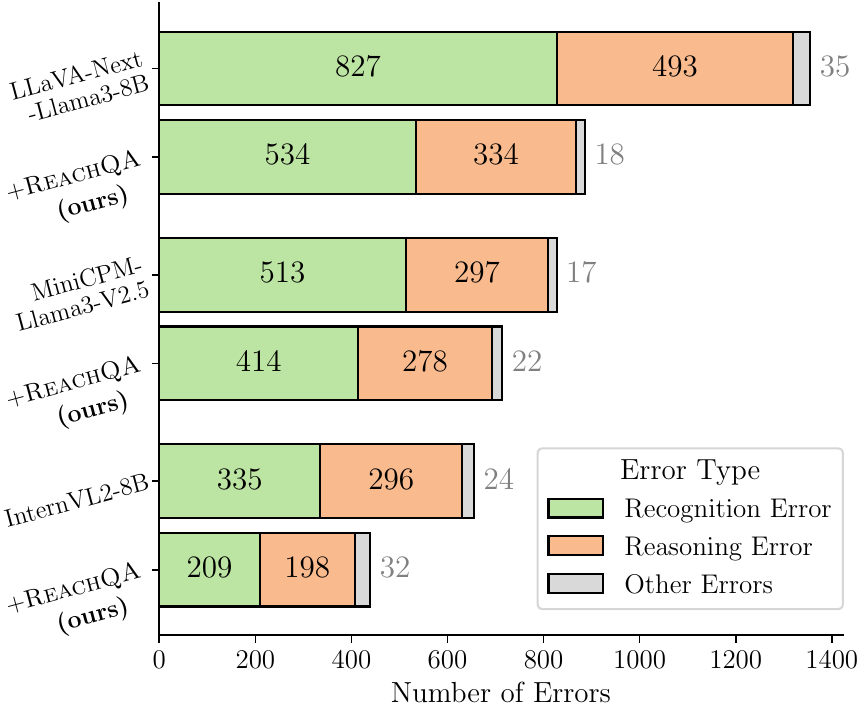} 
    \caption{Error distribution of three baseline models vs. our \data-trained versions on ChartQA test set \citep{chartqa}, as judged by GPT-4o.
    Error types are categorized into Recognition Error, Reasoning Error, and Other Errors (question misinterpretation, factual inconsistency or hallucination, and response refusal).}
    \label{fig:error_distribution}
    \end{center}
    \vspace{-15pt}
\end{figure}

Multimodal large language models (MLLMs) have achieved notable progress, particularly in visual recognition tasks \citep{gpt4o, claude3_5}. 
However, their ability to comprehend complex images like charts in real-world contexts and to address reasoning-intensive questions remains limited compared to humans \citep{chartqa, pixels, charxiv}.
Our analysis of the error distribution in ChartQA (Figure~\ref{fig:error_distribution}) also reveals two main failure modes in current MLLMs: while most errors originate from visual misrecognition, a substantial portion arises from flawed reasoning even when visual elements are correctly identified. 
This contrasts sharply with human performance \citep{mathv, charxiv}, since we can purposefully identify task-critical information from images and engage in step-by-step reasoning processes.
These observations motivate our investigation into bridging this capability gap through the acquisition of human-like reasoning patterns.

While distilling expert rationales from humans or stronger models presents a promising pathway for improving reasoning abilities \citep{chartllama, chartast, chartinstruct, chartgemma}, constructing high-quality training data for chart-related tasks is expensive and hard to scale.
Early approaches typically rely on manual chart collection from online sources, meticulous data filtering and annotation \citep{chartqa, charxiv}. 
Recent attempts to automate Q\&A generation through LLMs often use data tables as inputs \citep{chartllama, chartinstruct}, which neglect the visual-semantic features of charts.
Even with the use of MLLMs \citep{chartgemma}, our preliminary study (\textsection~\ref{pre_study}) shows they also struggle to produce accurate and challenging data for advanced reasoning skill acquisition.
In comparison, we find that when LLMs process charts in a better textual format---\textbf{code}, they can generate Q\&A pairs at lower costs and with higher quality.

Inspired by the concept of intermediary translation \citep{zarechnak1986intermediary, leon2007universal}, which refers to using a bridge language to improve translation quality across diverse languages in literary studies, we introduce \textbf{Code-as-Intermediary Translation (CIT)}. 
In this method, the code acts as an intermediary, converting chart images into textual representations by faithfully encoding visual-semantic features within itself. 
This process enables LLMs to understand cross-modal information more accurately, thereby generating visually complex Q\&A pairs with high-quality reasoning rationales.
Furthermore, it facilitates the adoption of text-based instruction augmentation strategies, such as Self-Instruct \citep{self_instruct} and Evol-Instruct \citep{evol_instruct}, to expand the quantity and enhance the complexity of the synthetic charts.
Starting with $33$ seed codes collected from the Matplotlib gallery, we synthesize more chart-plotting codes covering diverse types and topics, and then complicate them to create richer ones.
Finally, using the synthetic codes as a bridge, we generate charts (via Python) and instructions (via LLMs) in a bi-directional process, ensuring the alignment between modalities.

With the CIT method, we construct \textbf{\data}, a multimodal instruction dataset containing $3,249$ \textbf{rea}soning-intensive \textbf{ch}arts and $19,963$  Q\&A pairs, all at a remarkably low cost of just $\$300$.
The dataset comprises questions focused on both visual recognition and reasoning, designed to address the dual challenges of current MLLMs.
Additionally, we create a manually verified test set to assess models' recognition and reasoning abilities independently.
Experiments demonstrate that \data-trained models achieve substantial performance gains across benchmarks, with LLaVA-Next-Llama3-8B \citep{llavanext} improving by over $30\%$ on average, while both types of errors are significantly reduced (Figure~\ref{fig:error_distribution}).
Notably, these improvements generalize beyond chart-specific tasks to broader multimodal reasoning tasks like MathVista and MATH-Vision---an outcome previously unattainable with existing chart-focused datasets.
Finally, we explore \data{'s} working mechanism and more features, providing actionable guidelines for building performant multimodal datasets.

Our contributions are summarized as follows\footnote{ The code and datasets are now publicly available at \href{https://github.com/hewei2001/ReachQA}{https://github.com/hewei2001/ReachQA}.}:
\begin{enumerate}[leftmargin=*]
    \item We propose Code-as-Intermediary Translation (CIT), a cost-effective and efficient method for synthesizing multimodal instruction data with code as a bridge between the two modalities.
    \item Through CIT, we construct \data, the first fully LLM-synthesized reasoning-intensive chart Q\&A dataset, focusing on both visual recognition and reasoning abilities.
    \item We conduct extensive experiments and analyses to demonstrate \data{'s} effectiveness for MLLMs, along with its strong generalization to broader multimodal reasoning tasks.
\end{enumerate}

\begin{table*}[ht]
\centering
\resizebox{\linewidth}{!}{%
\begin{tabular}{@{}lcccccccccc@{}}
    \toprule[1.5pt]
    \multirow{3}{*}{\textbf{Datasets}} & \multicolumn{4}{c}{\textbf{Chart Properties}} & \multicolumn{3}{c}{\textbf{Q\&A Properties}} & \multicolumn{3}{c}{\textbf{Dataset Properties}} \\ 
    \cmidrule(lr){2-5} \cmidrule(lr){6-8} \cmidrule(lr){9-11} 
    & \makecell{\# Chart\\Type} & \makecell{\# Chart\\Topic} & \makecell{Textual\\Format} & \makecell{Vis.\\Comp.} & \makecell{Temp.\\Free} & \makecell{Vis.\\Refer.} & \makecell{Rat.\\Annot.} & \makecell{Train\\Set} & \makecell{Test\\Set} & \makecell{Scal.} \\ 
    \midrule
    PlotQA~\citep{plotqa}      & 3      & -  & Table   & \xmark & \xmark & \cmark & \xmark & \cmark & \cmark & \xmark \\
    ChartQA~\citep{chartqa}      & 3      & 15   & Table   & \xmark & \cmark & \cmark & \xmark & \cmark & \cmark & \xmark \\
    OpenCQA~\citep{opencqa}      & 5      & 10   & Caption & \xmark & \cmark & \xmark & \cmark & \xmark & \cmark & \xmark \\
    MathVista~\citep{MathVista}    & -    & -  & -     & \xmark & \cmark & \xmark & \xmark & \xmark & \cmark & \xmark \\
    CharXiv~\citep{charxiv}      & -    & -  & -     & \cmark & \notcheckmark & \cmark & \xmark & \xmark & \cmark & \xmark \\
    ChartBench~\citep{chartbench}   & \textbf{9 / 42} & -   & Table & \xmark & \xmark & \xmark & \xmark & \cmark & \cmark & \cmark \\
    ChartX~\citep{chartx}       & 18     & 22   & Code* & \xmark & \cmark & \xmark & \xmark &  \xmark & \cmark & \cmark \\
    \midrule
    MMC~\citep{mmc} & 6      & 5    & Caption & \cmark & \cmark & \xmark & \cmark & \cmark & \cmark & \notcheckmark \\
    ChartLlama~\citep{chartllama}   & 10     & -  & Table & \xmark & \cmark & \xmark & \cmark & \cmark & \cmark & \cmark \\
    ChartAst~\citep{chartast}     & 9      & -  & Table  & \xmark & \xmark  & \xmark & \cmark  & \cmark & \xmark & \notcheckmark \\
    ChartInstruct~\citep{chartinstruct} & -   & -  & Table   & \xmark & \cmark & \xmark & \cmark & \cmark & \xmark & \notcheckmark  \\
    ChartGemma~\citep{chartgemma}   & -    & -  & -     & \xmark & \cmark & \cmark & \cmark & \cmark & \xmark & \notcheckmark \\
    \textbf{\data (ours)}  & \textbf{10 / 32} & $\infty$ & \textbf{Code} & \cmark & \cmark & \cmark & \cmark & \cmark & \cmark & \cmark  \\
    \bottomrule[1.5pt]
\end{tabular}}
\caption{Comparison of existing chart-related datasets.
Only the chart Q\&A task is considered, though some datasets include multiple tasks.
Abbreviations: Vis.=visual, Comp.=complexity, Temp.=template, Refer.=Reference, Rat.=rationale, Annot.=annotation and Scal.=scalable.
Cells marked with ``\notcheckmark'' indicate mixed attributes (e.g., partially template-based; scalable Q\&A but non-scalable chart data.).
``/'' means the dataset includes multiple chart type granularity.
``*'' indicates while chart-plotting codes are public, their Q\&A synthesis still relies on data tables.
}
\label{tab:compare}
\vspace{-8pt}
\end{table*}

\section{Background}

\subsection{Deficiencies in Existing Chart Datasets}

Existing chart-related datasets are either collected from online data sources or generated by models, sometimes requiring manual annotation or automated question generation. 
Most of them focus on visual recognition tasks. 
While some recent works target advanced reasoning, they often struggle with scalability or other shortcomings.
Table~\ref{tab:compare} summarizes these datasets, with further details below.

\paragraph{Chart Properties.}
The \textit{visual diversity} is shaped by the variety of chart types and topics \citep{charxiv}.
Early datasets like ChartQA~\citep{chartqa} and OpenCQA~\citep{opencqa}, sourced from limited websites, featured uniform styles with minimal diversity.
To address this, recent works like ChartAst~\citep{chartast} synthesize charts with randomized attributes (e.g., color, fonts) using LLMs.
However, beyond the superficial variations in chart appearance, many of them overlook the \textit{visual complexity} \citep{advancing}.
As models evolve, simple style changes no longer pose challenges.
Datasets like CharXiv~\citep{charxiv} and MMC~\citep{mmc}, which include complex scientific charts from arXiv papers, naturally exhibit greater complexity in recognition.
Additionally, the textual format of charts is critical, enabling dataset expansion via language models.

\paragraph{Q\&A Properties.}
Some benchmarks like PlotQA~\citep{plotqa} and ChartBench~\citep{chartbench} use predefined templates to generate Q\&A pairs, resulting in monotonous and simplistic questions.
Other datasets, such as ChartQA~\citep{chartqa} and CharXiv~\citep{charxiv}, required manual annotation, which improved quality but increased costs and hindered scalability.
With the advent of LLMs, works like ChartLlama~\citep{chartllama} and ChartInstruct~\citep{chartinstruct} use them to generate diverse questions from data tables while also providing rationale annotations for training.
However, these methods fail to capture visual elements like color, layout, and structure because they rely on only the data table.
Thus, the generated Q\&A pairs lack \textit{visual references}, undermining the inherently multimodal nature of this task.
To address this, ChartGemma~\citep{chartgemma} uses MLLMs to generate Q\&A directly from charts.

\paragraph{Dataset Properties.}
While manually annotated datasets like MathVista~\citep{MathVista} and CharXiv~\citep{charxiv} provide high-quality data, their development is resource-intensive, typically resulting in datasets of only a few thousand samples.
In the era of LLMs, such methods are impractical for scaling to the size needed to train larger models.
Recent efforts, such as ChartAst~\citep{chartast}, ChartInstruct~\citep{chartinstruct}, and ChartGemma~\citep{chartgemma}, have explored Q\&A generation for dataset expansion, but they remain limited by the difficulty of collecting a large set of charts.
A more scalable approach is to leverage the generative capabilities of LLMs to synthesize charts like ChartBench~\citep{chartbench} and ChartX~\citep{chartx}.

\begin{figure*}[ht]
    \begin{center}
    \includegraphics[width=0.96\textwidth]
    {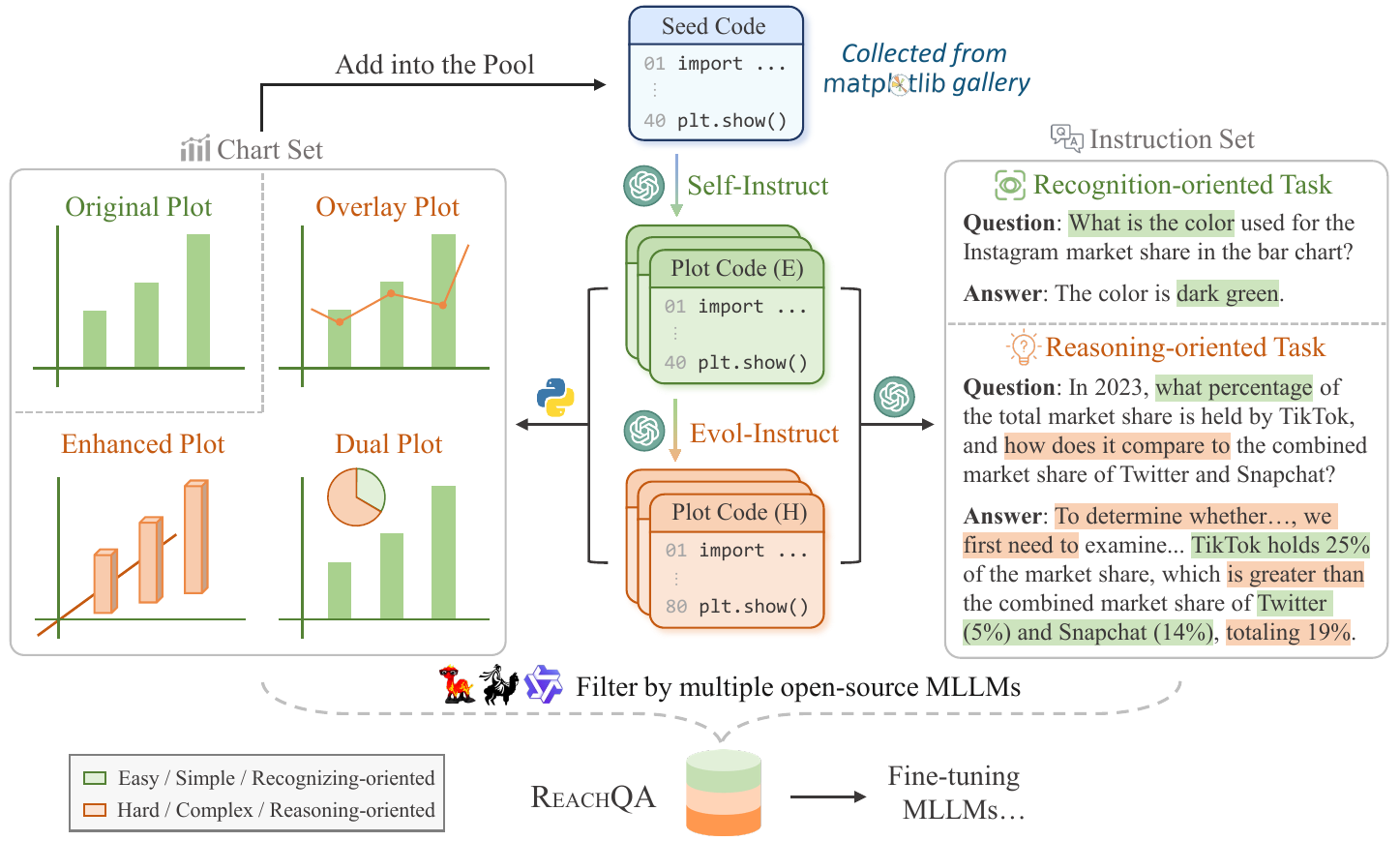} 
    \caption{Overview of the Code-as-Intermediary Translation (CIT) method for synthesizing multimodal instruction data. 
    The process starts with 33 seed codes, generating plot codes across various chart types, topics, and complexity levels via Self-Instruct and Evol-Instruct. 
    The chart and instruction sets are constructed bi-directionally, and the final filtering yields \data, a dataset for distilling visual chart reasoning abilities from LLMs to MLLMs.}
    \label{fig:reachqa}
    \end{center}
    \vspace{-12pt}
\end{figure*}

\subsection{Can LLMs Understand Charts without Visual Input?} \label{pre_study}
To explore whether there is a more effective textual format for representing visual information than data tables, we propose using \textbf{code}. 
By precisely encoding chart structures and details, the code may serve as an ideal bridge between modalities. 
We design an experiment to test this hypothesis.
We first collect 25 complex charts, along with their corresponding data tables and code, from authentic research papers. 
These charts often feature multiple or overlay plots and dense data groups, with the code averaging over 100 lines.
For each sample, GPT-4o receives three types of input—table, code, and chart images—to generate a challenging Q\&A pair.
In total, 75 pairs are created, randomly shuffled, and then presented to annotators for blind evaluation.
The annotators are asked to rate each pair on accuracy, reasoning complexity, and visual reference, using a scale of 1 (low) to 3 (high). 

\begin{table}[t]
\centering
\vspace{2pt}
\resizebox{0.38\textwidth}{!}{%
\begin{tabular}{lcccc}
\toprule[1.5pt]
Input & Acc. & \makecell{Reas.\\Comp.} & \makecell{Vis.\\Refer.} & Cost (\$) \\ 
\midrule
Table & \textbf{2.72} & 2.51 & 1.19 & \textbf{0.047} \\
Code  & 2.60 & \textbf{2.56} & 2.15 & 0.092 \\
Chart & 1.91 & 1.53 & \textbf{2.36} & 0.107 \\
\bottomrule[1.5pt]
\end{tabular}
}
\caption{Rating results for three input types in our study.}
\label{tab:user}
\vspace{-10pt}
\end{table}

The results in Table \ref{tab:user} indicate that both text-based inputs outperform visual chart input in the first two aspects, with code scoring 2.60 in accuracy (vs. 1.91) and 2.56 in reasoning complexity (vs. 1.53). 
As expected, table input has the lowest visual reference score (1.19), while chart input scores highest in this (2.36), confirming the ability of MLLMs to directly interpret visual information.
Surprisingly, despite the absence of visual input to the model, the code achieves a relatively high visual reference score (2.15), highlighting its potential to translate chart images into textual representations.

\section{Methodology}
Building on the findings above, we propose Code-as-Intermediary Translation (CIT), a data synthesis method for distilling visual reasoning abilities from LLMs to MLLMs, as illustrated in Figure~\ref{fig:reachqa}.
In the following sections, we describe how we synthesize intermediate codes (\textsection~\ref{sys_code}), generate paired charts and instructions (\textsection~\ref{sys_chart_qa}), ensure data quality (\textsection~\ref{quality}), and ultimately construct our dataset, \data.

\subsection{Intermediary Code Synthesis} \label{sys_code}

\paragraph{Seed Code Collection.}
We start by collecting a small set of $33$ seed code samples, which we refer to as $C_{\text{seed}}$.
These samples are sourced directly from the official Matplotlib gallery\footnote{https://matplotlib.org/stable/gallery/index.html} to ensure quality and minimize manual effort. 
Collectively, these code samples cover a diverse range of chart types, including common types like bar, line, and scatter charts, as well as more specialized charts such as bubble, contour, and donut charts.
All samples are verified for executability to guarantee the reliability of the subsequent code synthesis process.

\paragraph{Self-Instruct for Diverse Code Generation.} 
To expand the diversity and coverage of the chart set, we apply the Self-Instruct method \citep{self_instruct}, which synthesizes instruction data by prompting LLMs with existing ones as few-shot examples \citep{few_shot}.
In our approach, we provide $3$ randomly selected code snippets as examples in each generation, guiding the model to synthesize chart-plotting code of the same kind.

To diversify chart generation, a chart type is randomly chosen from 10 major and 32 minor categories for the model to generate.  
For chart content, we provide two topic options, allowing the model to freely combine or expand on these themes based on its knowledge, leading to varied topics and data.
A chain-of-thought (CoT) process \citep{cot} is used for code generation, starting with the chart's background and data, followed by the final executable code. 
This step-by-step approach ensures logical coherence and code functionality. 
The generated codes are referred to as $C_{\text{easy}}$ for use in subsequent phases of the construction.
The chart types and topics are detailed in Appendix~\ref{app:type_topic}.

\paragraph{Evol-Instruct for Complex Code Generation.} 
To enhance the visual complexity of the synthetic charts, we adopt the Evol-Instruct method \citep{evol_instruct}, which leverages LLMs to evolve simple chart-plotting code into more complex versions by presenting existing code alongside an evolution strategy as context.
It addresses a key limitation in prior work that emphasizes the quantity of charts while often neglecting the difficulty of chart interpretation.
Starting with code samples from $C_{\text{easy}}$, we apply one of the following predefined evolution directions:
(1) expanding the data size or number of data groups; (2) adding or modifying visual elements to enhance presentation; (3) overlaying a different type of chart on the original plot; (4) introducing an additional subplot beside the original plot.
These strategies ensure that the resulting charts demand more nuanced visual interpretation and in-depth reasoning.
We follow a CoT process like previous steps, where the model first analyzes the existing code and then generates the evolved one. 
The evolved codes, referred to as $C_{\text{hard}}$, are also added to the code pool for subsequent use.

\subsection{Bi-directional Translation} \label{sys_chart_qa}

\paragraph{Chart Generation through Code Execution and Self-Repair.} 
We generate charts by executing all the Python plotting code. 
However, during the generation and evolution process, program errors are inevitable.
To ensure correctness, we will validate the code before adding it to the pool. 
When errors occur, the code is not immediately discarded; instead, we apply a Self-Repair method \citep{self_debug}, feeding the code and execution results into the LLMs for correction.
This process repeats until the code is fixed or reaches an iteration limit, after which the code is discarded if it remains faulty.
On average, this approach fixes about 15\% of the code generated by GPT-4o, with 5\% remaining unrepairable and filtered out, yielding $C_{\text{final}}$.

\paragraph{Instruction Generation through Guided Prompting.}

After verifying executability, we use $C_{\text{final}}$ to create instruction sets in the form of Q\&A pairs.
Building on prior work of in-context Q\&A generation \citep{self_icl, self_demos}, we guide the model in two steps: first generating a batch of questions, then producing corresponding answers.
To ensure high-quality answers, we also employ a step-by-step approach where the model first provides detailed calculations and analyses, which are then refined into concise, educational answers optimized for learning \citep{textbooks}.
The model generates two types of instructions: \textit{recognition-oriented}, focused on visual information retrieval, and \textit{reasoning-oriented}, requiring both recognition and multi-step reasoning. 
With minimal constraints on content, the model is encouraged to explore creative and diverse instructions. 
Multiple questions can be generated for each chart, and redundant ones are filtered using ROUGE-L overlap, following \citet{self_instruct}.

\subsection{Quality Assurance} \label{quality}

\paragraph{Multimodal Validation for Enhanced Data Quality.}
Although our dataset is fully synthesized using LLMs, we acknowledge the importance of integrating visual information to enhance data quality \citep{chartgemma, advancing}. 
Thus we introduce a multimodal validation step, using MLLMs to verify both generated charts and their corresponding instructions.
Since models differ in architecture, visual encoders, and training recipes, they may focus on varying aspects of the images. 
Taking this into account, we adopt a ``majority voting'' approach by ensembling multiple smaller, locally hosted models.
This ensures reliable visual validation while remaining cost-effective.
For chart validation, each model rates charts on a scale of 1 to 5, and those below a threshold are filtered out.
For instructions, both Q\&A pairs and corresponding charts are fed into the models and verified, with multiple negative votes leading to sample rejection. 

\paragraph{Testing Set Construction and Annotation Refinement.}
For the \data testing set, we follow a similar process as in previous data generation but apply stricter filtering criteria to ensure higher quality. 
Additional annotators are recruited for manual review and refinement. For the charts, they first check the images to identify any potential visual errors. 
For the Q\&A pairs, they ensure the questions are relevant to the chart and answerable, then correct any hallucinations or logical inconsistencies in the answers.
Afterwards, two rounds of review are conducted to confirm the questions meet the multimodal recognition or reasoning standards in our settings.
Only samples with agreement from at least two annotators are included. 
The inter-annotator agreement, with a kappa coefficient of $0.82$, indicates strong consistency \citep{kappa}. 
Table \ref{tab:stat} presents the final dataset statistics.

\begin{table}[t]
\centering
\setlength{\tabcolsep}{3pt}
\resizebox{0.45\textwidth}{!}{%
\begin{tabular}{lrr}
\toprule[1.5pt]
\textbf{Statistics} & \textbf{Train Set} & \textbf{Test Set} \\
\midrule
Total charts & 3,249 & 500 \\
- \# Chart types & 10 / 32 & 10 / 32 \\
- \# Overlay plots & 1,030 & 220 \\
- \# Multiple plots & 593  & 251 \\
- Average size (px) & 2480$\times$1571 & 2798$\times$1601 \\
\midrule
Unique questions & 19, 963 & 2,000 \\
- \# Reco. per chart & 2.53 & 2 \\
- \# Reas. per chart & 3.62 & 2 \\
\midrule
Avg. Reco. Q. length & 22.1 & 21.0 \\
Avg. Reco. A. length & 38.3 & 7.0 \\
Avg. Reas. Q. length & 38.2 & 35.4 \\
Avg. Reas. A. length & 68.4 & 24.9 \\

\bottomrule[1.5pt]
\end{tabular}
}
\caption{\data dataset statistics. Sequence lengths are calculated based on the GPT-4o tokenizer.}
\label{tab:stat}
\vspace{-10pt}
\end{table}

The total cost of data construction, excluding open-source model usage and annotation labor for the testing set, was about $\text{\$}300$. The detailed expense breakdown is provided in Appendix~\ref{app:cost}.
Since all data splits are generated using the same process and model, we analyze potential data contamination in Appendix~\ref{app:contamination}.
The prompt templates we use in each step are shown in Appendix~\ref{app:prompt}.

\section{Experiments}

\subsection{Experimental Setups} \label{sec:setup}

\paragraph{Benchmarks.}
We evaluate the models on three categories of tasks that cover both chart-related and general multimodal recognition and reasoning.
First, traditional chart-related benchmarks are considered, including ChartQA, ChartBench, and ChartX, which primarily test recognition capabilities. 
Second, we assess novel chart-related benchmarks that require both recognition and reasoning, including CharXiv and our \data test set. 
Third, we evaluate general multimodal reasoning abilities on MathVista and MATH-Vision.

\paragraph{Models and baselines.}
We evaluate a range of MLLMs from three categories:
(1) Powerful proprietary models, including GPT-4o \citep{gpt4o}, GPT-4o mini \citep{gpt4o_mini}, and Claude 3.5 Sonnet \citep{claude3_5}.
(2) Chart-augmented open-source models, such as ChartInstruct-7B \citep{chartinstruct}, ChartAssistant-13B \citep{chartast}, and ChartGemma-3B \citep{chartgemma}, which are specifically enhanced for chart-related tasks.
(3) Latest general open-source models, including LLaVA-Next-Llama3-8B \citep{llavanext}, MiniCPM-Llama3-V2.5-8B \citep{minicpm}, and InternVL2-8B \citep{internvl}. 
For each general model, we conduct supervised fine-tuning (SFT) using the \data training set.
Specifically, we train three variants: one using 8k recognition-oriented samples (denoted as Reco.), one using 12k reasoning-oriented samples (denoted as Reas.), and a combined version incorporating both  (denoted as All).
More details on the datasets and evaluation can be found in Appendix~\ref{app:exp}.

\subsection{Experimental Results} \label{exp:main}

\begin{table*}[t]
\belowrulesep=0pt
\aboverulesep=0pt
\fontsize{14}{21}\selectfont
\setlength{\tabcolsep}{4pt}
\centering

\resizebox{\textwidth}{!}{%
\begin{tabular}{lc|cccc|cccc|cccc}
\toprule[1.5pt]
\multirow{2}{*}{\textbf{Models}} & \multirow{2}{*}{\textbf{Avg. \color{darkgreen}{($\uparrow$)}}} & \textbf{ChartQA} & \multicolumn{2}{c}{\textbf{ChartBench}} & \textbf{ChartX} & \multicolumn{2}{c}{\textbf{\data}} & \multicolumn{2}{c|}{\textbf{CharXiv}} & \multicolumn{2}{c}{\textbf{MathVista}} & \textbf{MATH-V} \\
\cmidrule(lr){3-3} \cmidrule(lr){4-5} \cmidrule(lr){6-6} \cmidrule(lr){7-8} \cmidrule(lr){9-10} \cmidrule(lr){11-12} \cmidrule(lr){13-13}
& & \textbf{QA} & \textbf{Binary} & \textbf{NQA} & \textbf{QA} & \textbf{Reas.} & \textbf{Reco.} & \textbf{Reas.} & \textbf{Desc.} & \textbf{Math} & \textbf{General} & \textbf{QA} \\
\midrule
\multicolumn{13}{c}{\textbf{Proprietary Multimodal Large Language Models}} \\
\midrule
GPT-4o mini & 49.34 & 77.52 & 70.26 & 34.93 & 35.45 & 27.20 & 53.50 & 34.10 & 74.92 & \multicolumn{2}{c}{56.70} & 28.85 \\
GPT-4o & 59.85 & 85.70 & \textbf{81.03} & \textbf{52.88} & 46.60 & 39.70 & 66.80 & 47.10 & \textbf{84.45} & \multicolumn{2}{c}{63.80} & 30.39 \\
Claude 3.5 Sonnet & \textbf{64.50} & \textbf{90.80} & 76.72 & 48.29 & \textbf{58.24} & \textbf{51.70} & \textbf{74.30} & \textbf{60.20} & 84.30 & \multicolumn{2}{c}{\textbf{67.70}} & \textbf{32.76} \\
\midrule
\multicolumn{13}{c}{\textbf{Chart-augmented Multimodal Large Language Models}} \\
\midrule
ChartInstruct-7B & 25.93 & 66.64 & 61.40 & 26.95 & 26.62 & 6.00 & 10.50 & 8.80 & \textbf{21.40} & 15.37 & 31.52 & \textbf{10.07} \\
ChartAssistant-13B & 28.25 & 79.90 & 58.15 & 24.62 & 23.20 & \textbf{10.70} & 19.60 & 11.70 & 16.93 & 17.78 & \textbf{39.57} & 8.55 \\
ChartGemma-3B & \textbf{33.08} & \textbf{80.16} & \textbf{78.90} & \textbf{34.10} & \textbf{35.15} & 9.20 & \textbf{27.80} & \textbf{12.50} & 21.30 & \textbf{19.07} & 38.04 & 7.70 \\
\midrule
\multicolumn{13}{c}{\textbf{Open-Source Multimodal Large Language Models}} \\
\midrule
LLaVA-Next-Llama3-8B & 24.46 & 45.80 & 42.90 & 15.86 & 15.45 & 6.50 & 17.90 & 17.20 & 31.45 & 22.41 & 44.13 & 9.44 \\
\ \ + \data (Reco.) & 32.88 \color{darkgreen}{(+34.4\%)} & \textbf{66.96} & 56.95 & \textbf{29.52} & \textbf{27.25} & 8.80 & 29.00 & 22.20 & 32.58 & 27.40 & 49.78 & 11.25 \\
\ \ + \data (Reas.) & 32.39 \color{darkgreen}{(+32.4\%)} & 64.48 & 56.80 & 25.14 & 25.90 & 8.40 & 26.30 & \textbf{22.70} & \textbf{35.67} & \textbf{28.89} & \textbf{50.65} & \textbf{11.38} \\
\ \ + \data (All) & \textbf{32.98 \color{darkgreen}{(+34.8\%)}} & 64.56 & \textbf{57.00} & 29.33 & 27.08 & \textbf{11.10} & \textbf{29.60} & 22.50 & 32.33 & 27.59 & 50.43 & 11.25 \\
\cdashline{1-13}
MiniCPM-Llama3-V2.5 & 33.39 & 66.92 & 48.90 & 22.29 & 23.72 & 10.30 & 25.30 & 22.00 & 46.20 & 37.22 & 53.04 & 11.45 \\
\ \ + \data (Reco.) & 38.62 \color{darkgreen}{(+15.7\%)} & 71.12 & \textbf{56.65} & \textbf{33.29} & 29.53 & 10.60 & 34.10 & 25.60 & \textbf{48.75} & 41.48 & \textbf{60.43} & 13.22 \\
\ \ + \data (Reas.) & 38.52 \color{darkgreen}{(+15.4\%)} & 71.72 & \textbf{56.65} & 29.62 & 28.23 & \textbf{11.00} & 33.00 & 27.50 & 48.70 & \textbf{43.52} & 60.22 & 13.52 \\
\ \ + \data (All) & \textbf{38.67 \color{darkgreen}{(+15.8\%)}} & \textbf{71.44} & 55.80 & 30.43 & \textbf{29.68} & \textbf{11.00} & \textbf{35.10} & \textbf{28.30} & 47.62 & 42.22 & 60.00 & \textbf{13.75} \\
\cdashline{1-13}
InternVL2-8B & 40.03 & 73.80 & 52.05 & 32.86 & 35.10 & 16.20 & 33.70 & 26.30 & 46.10 & 46.11 & 61.74 & 16.38 \\
\ \ + \data (Reco.) & 48.21 \color{darkgreen}{(+20.4\%)} & \textbf{82.92} & \textbf{66.35} & 46.14 & \textbf{46.62} & 19.90 & 49.50 & 32.20 & 54.38 & 47.96 & \textbf{67.61} & 16.78 \\
\ \ + \data (Reas.) & 47.87 \color{darkgreen}{(+19.6\%)} & 82.84 & 64.05 & 46.52 & 44.88 & 20.10 & 49.40 & \textbf{32.80} & 52.40 & \textbf{49.44} & 66.52 & \textbf{17.66} \\
\ \ + \data (All) & \textbf{48.35 \color{darkgreen}{(+20.8\%)}} & 82.44 & 65.90 & \textbf{47.29} & 45.38 & \textbf{21.30} & \textbf{49.80} & 32.70 & \textbf{54.83} & 48.89 & 66.30 & 17.01 \\
\bottomrule[1.5pt]
\end{tabular}%
}
\caption{Evaluation results on seven benchmarks.
The best performance for each category and task is in \textbf{bold}. 
The percentage of performance improvements compared to the vanilla model is denoted by {\color{darkgreen}{($\uparrow$)}}. 
}
\label{tab:main}
\vspace{-10pt}
\end{table*}

Table~\ref{tab:main} presents the quantitative results for all models across each task.
We can find that:

\paragraph{Synthetic datasets can also effectively measure abilities.}
Our \data test set effectively evaluates models' reasoning and recognition skills, showing trends similar to human-annotated datasets like CharXiv. 
For instance, GPT-4o exhibits a reasoning score of 39.70 and a recognition score of 66.80 on \data, closely mirroring its performance on CharXiv (i.e., 47.10 and 84.45, respectively). 
This consistency suggests that LLM-generated datasets, with minimal human intervention, can rival human-labeled data. 
Moreover, \data presents a significant challenge to models' visual abilities, as random guessing results in very low scores. 
In contrast, traditional benchmarks like ChartQA may allow models to leverage pre-existing knowledge, inflating results without truly testing visual capabilities \citep{mmmupro}.

\paragraph{Proprietary models demonstrate more balanced performance.}
Proprietary models like GPT-4o achieve competitive results on both traditional chart-related tasks and reasoning-intensive tasks like \data and CharXiv.
In contrast, open-source models, whether chart-augmented or general-purpose, excel in recognition tasks but struggle in complex ones. 
This disparity highlights their imbalanced capabilities, and also suggests potential overfitting to simpler charts.
Although proprietary models may not always lead in specific tasks, the stable and balanced performance makes them more suitable for real-world applications.

\paragraph{Specialized training data significantly improves model performance.} 
Models trained on 8k \data recognition data outperform in recognition tasks, while those trained on 12k reasoning data could do better in reasoning tasks. 
When both data types are combined (i.e., 20k in total), models see the greatest improvement, with performance increasing by at least 15\% across all models we test. 
Notably, the LLaVA-Next-Llama3-8B achieves a 34.8\% boost in average performance.
This suggests that a model’s visual capability comprises two complementary aspects, and training on both data types together produces optimal results.
Moreover, despite the absence of math-target data in the training set, the models generalize well to the MathVista and MATH-Vision benchmarks, highlighting the transferability of multimodal reasoning abilities distilled from expert rationales.

\section{Discussion}

\subsection{The Role of Expert Rationales}

\begin{table}[t]
\centering
\vspace{4pt}
\setlength{\tabcolsep}{3pt}
\resizebox{0.48\textwidth}{!}{%
\begin{tabular}{lccccc}
    \toprule[1.5pt]
    \textbf{Models}               & \textbf{Avg.}  & \textbf{\data} & \textbf{CharXiv} & \textbf{MathVista} & \textbf{Math-V} \\
    \midrule
    Base Model      & 16.39     & 6.50          & 17.20         & 32.40          & 9.44       \\
    + ChartBench    & 17.06     & 7.30          & 17.00         & 33.60          & 10.33      \\
     + ChartAst     & 17.67     & 7.10          & \underline{20.40}  & 32.10    & \underline{11.08}  \\
    + The Cauldron  & 18.61     & \underline{10.10} & 19.10     & 35.60 & 9.64 \\
    + ChartGemma    & \underline{19.11}         & 10.00         & 19.40    & \underline{36.40}  & 10.62      \\
     + \data        & \textbf{20.74} & \textbf{11.10} & \textbf{22.50}   & \textbf{38.10} & \textbf{11.25}  \\
    \bottomrule[1.5pt]
\end{tabular}
}
\caption{Performance comparison of models trained on different datasets. The \data and CharXiv scores refer to reasoning splits here.}
\label{tab:generalization_ability}
\vspace{-6pt}
\end{table}

We analyze how training data quality affects visual reasoning abilities by comparing major open-source datasets (ChartBench, ChartAst, ChartGemma and The Cauldron\footnote{Unlike other datasets, The Cauldron \citep{the_cauldron} is selected for its generality as a collection of 50 vision-language datasets, from which we use 7 chart-related subsets.}).
We uniformly sample 20k Q\&A instructions from each dataset and train LLaVA-Next-Llama3-8B under controlled settings.

As shown in Table \ref{tab:generalization_ability}, the model trained on ChartBench performs the worst, likely due to the absence of reasoning steps in its responses. 
Although ChartAst includes rationale annotations, the template-based questions limit its effectiveness for learning reasoning patterns.
The model trained on the mixed dataset of The Cauldron show modest improvements, but is still restricted by the subsets' quality.
In contrast, models trained on ChartGemma and \data perform better, likely due to the distillation of expert rationales from stronger models (e.g., Gemini Flash 1.5 and GPT-4o), which directly affect the visual reasoning abilities. 
Additionally, we believe the visual richness of charts, as detailed in Appendix~\ref{app:train_data}, may also help improve generalization.

\subsection{Interaction between Recognition \& Reasoning Abilities} \label{sec:Interaction}

\begin{figure}[t]
    \centering
    \includegraphics[width=0.46\textwidth]{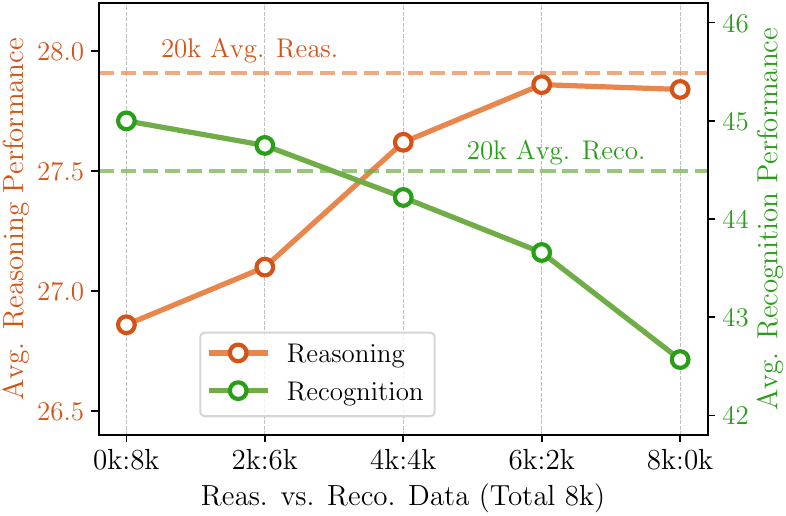}
    \caption{Performance comparison of different training data ratios with 8k total data. The dashed line represents the model's performance trained with full 20k data.}
    \label{fig:line}
    \vspace{-10pt}
\end{figure}

As previously noted, the recognition and reasoning abilities are likely interdependent. 
\citet{charxiv} suggest that recognition skills serve as prerequisites for effective reasoning.
To investigate this further, we conduct an experiment with LLaVA-Next-Llama3-8B, using a fixed 8k total training data size and varying the ratio of reasoning to recognition data from 0:8 to 8:0. 
We evaluated the models on recognition tasks (i.e., ChartQA, ChartBench, ChartX) and reasoning tasks (i.e., \data-Reas., CharXiv-Reas., MathVista).

Figure \ref{fig:line} shows that increasing the proportion of recognition or reasoning data improves performance on the respective tasks.
Models with more recognition data outperform those trained on 20k full data for recognition tasks. 
However, the reasoning performance gains plateau and even decline when reasoning data exceeds 50\%, suggesting diminishing returns when reasoning data is overemphasized. 
This supports the hypothesis that reasoning abilities are partially dependent on recognition skills.
When the model fails to interpret the image accurately, its reasoning ability is likely compromised \citep{charxiv}. 
Although this study is limited by data constraints, we expect the interaction between recognition and reasoning to become more pronounced with larger datasets.

\subsection{Balancing General \& Specialized Abilities} \label{sec:balancing}

\begin{figure}[t]
    \centering
    \includegraphics[width=0.44\textwidth]{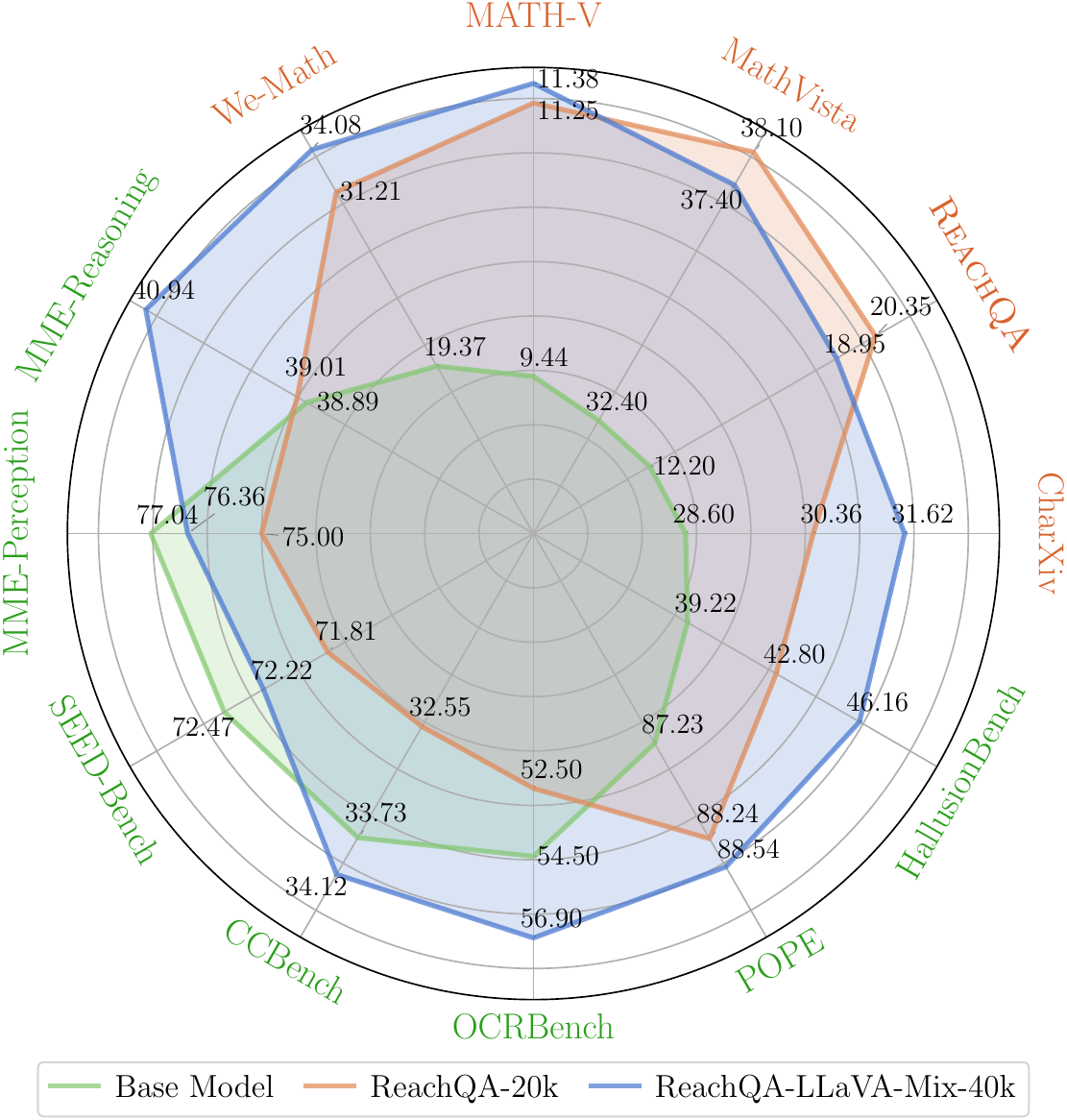}
    \caption{Performance comparison of models on 7 \textcolor{Green}{general} tasks and 5 \textcolor{Orange}{specialized reasoning} tasks.}
    \label{fig:radar}
    \vspace{-15pt}
\end{figure}

We investigate how models trained on specialized data perform on general-purpose multimodal tasks. 
Using 7 general multimodal benchmarks and 5 reasoning-focused benchmarks, we test three versions of LLaVA-Next-Llama3-8B: the vanilla model, the one trained on 20k \data samples, and another trained on 20k \data plus 20k general-purpose multimodal data sampled from 779k LLaVA-NeXT-Data\footnote{https://huggingface.co/datasets/lmms-lab/LLaVA-NeXT-Data}.
This dataset is chosen because the LLaVA-NeXT family of models was officially fine-tuned on it \citep{llavanext}, allowing us to approximate its original data distribution.

As shown in Figure \ref{fig:radar}, the vanilla model (green area) struggles with reasoning tasks, while the \data-trained model (orange area) improves on reasoning but loses general performance. 
Surprisingly, by incorporating only 20k general data (blue area), the model not only restores its general multimodal performance but also retains the enhanced reasoning ability.
It results in a well-balanced model with notable reasoning improvements and minimal drops in general domains.

\subsection{A Qualitative Analysis: From the Perspective of Attention Mechanism} \label{app:attention}

\begin{figure*}[ht]
    \begin{center}
    \includegraphics[width=0.99\textwidth]
    {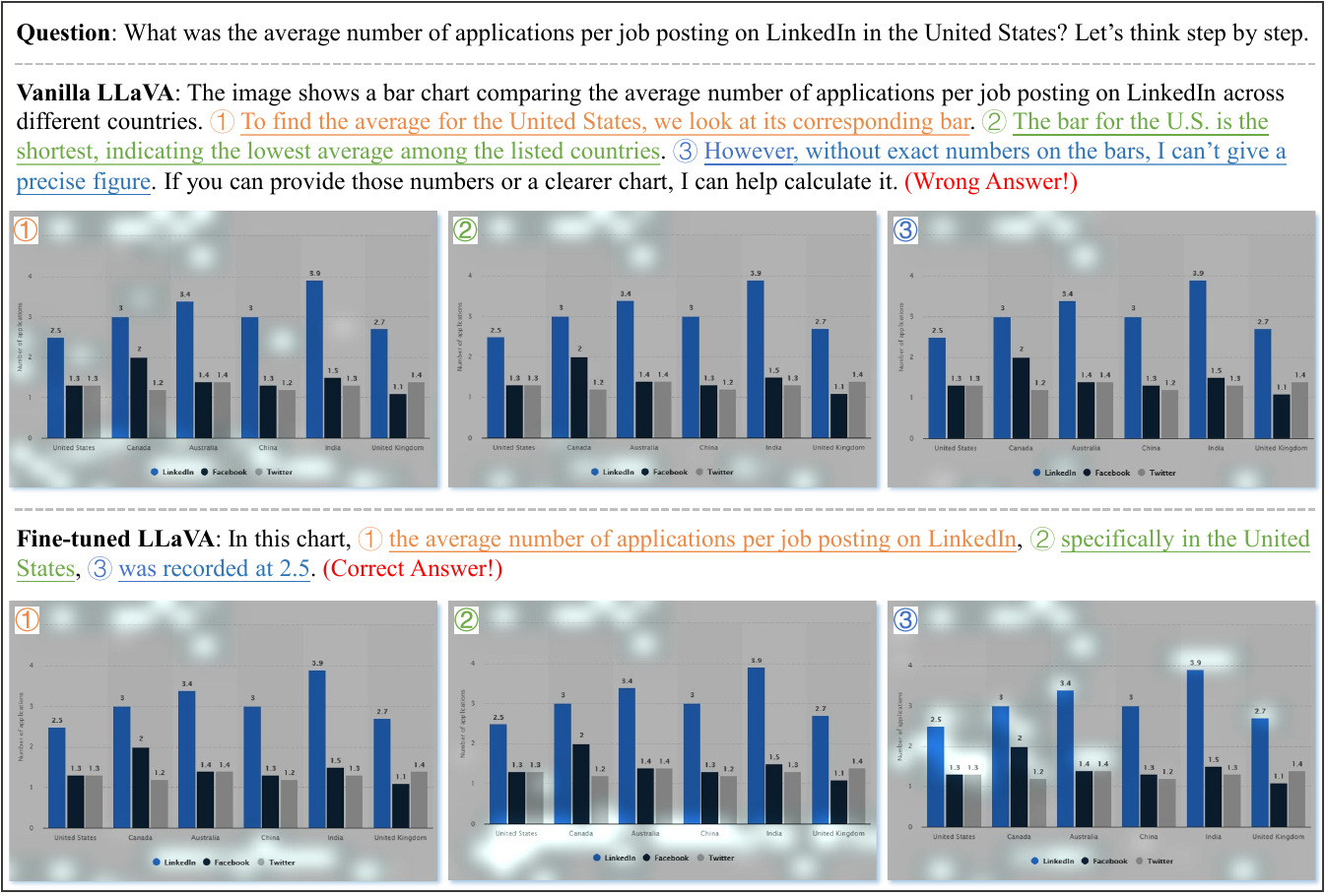} 
    \caption{An example of \textbf{attention visualization} from the ChartQA dataset. The top row shows the results from the vanilla LLaVA-Next-Llama3-8B model, while the bottom row displays the results from our fine-tuned model. For each output, we present the attention distribution (highlighted zones) at \textbf{three key steps}, calculated by averaging the attention values of all tokens in each step.}
    \label{fig:attention}
    \end{center}
    \vspace{-10pt}
\end{figure*}

To explore the mechanism behind the improved performance of our fine-tuned model, we conduct an analysis of the attention patterns during the next token prediction \citep{evit, colpali}. 
Figure \ref{fig:attention} presents a comparative case study between the vanilla model and the fine-tuned model.
Here, we apply full-parameter fine-tuning instead of LoRA to induce more pronounced changes in the attention layers \citep{lora}. 
The results show that the vanilla model produces lengthy outputs with redundant analysis and dispersed attention across the image, reaching a wrong conclusion at the end.
In contrast, the fine-tuned model identifies the key information at each step, with attention that accurately focuses on relevant visual elements (i.e., labels, axes and values).

This suggests that the model not only imitates expert rationales but also learns the underlying attention patterns crucial for effective visual reasoning. 
The model automatically establishes a synergistic relationship between recognition and reasoning capabilities, understanding what to recognize during the reasoning process and utilizing these recognition results to guide subsequent reasoning steps.

\section{Conclusion}
In this work, we delve into the key challenges MLLMs face in complex chart Q\&A tasks, highlighting their deficiencies in both recognition and reasoning. 
Building on our analysis of existing datasets and the untapped potential of LLMs, we propose Code-as-Intermediary Translation (CIT) as a novel method for distilling LLMs' abilities to improve MLLMs.
With code as a bridge between visual and textual modalities, CIT enables language models to interpret complex charts more precisely, facilitating the generation of higher-quality Q\&A pairs.
Our synthetic dataset, \data, demonstrates significant performance improvements across multiple models and benchmarks, with gains extending beyond chart-specific tasks to broader multimodal reasoning.
We believe CIT offers a promising direction for scalable and cost-effective multimodal instruction data synthesis.

\section*{Limitations}
We summarize the limitations of our method as follows:
(1) While CIT effectively uses code to link text and abstract images like charts and diagrams, applying this approach to natural images remains challenging. 
Current text-to-image models still lack precise control over fine details \citep{dalle, controlnet}, which can lead to misaligned synthetic data. 
Once more controllable techniques are developed, the synthesis of multimodal data could become more flexible and applicable.
(2) Although multimodal validation steps were introduced to reduce errors, the synthesized charts and Q\&A pairs might still contain occasional inaccuracies. 
Therefore, to ensure data quality for larger-scale applications, stronger models and stricter thresholds are essential.
(3) Our method may not be as effective for teacher models with limited capabilities, as it is inherently on a form of distillation \citep{distill_hinton}. 
The success of distillation depends on the strength of the teacher model, and in our scenario, weaker models may face challenges in interpreting charts via code.
Nevertheless, we believe that future models will not only become more capable but also more cost-efficient for data synthesis.

\section*{Ethical Considerations}
This research utilizes synthetic datasets for experimentation. 
We have ensured that all datasets comply with relevant ethical and privacy standards. 
All synthetic data have been rigorously processed to prevent the disclosure of any potentially sensitive information. 
We are committed to adhering to the ACL's ethical policies, ensuring transparency and reproducibility throughout the research process.

\section*{Acknowledgement}
The authors wish to thank the anonymous reviewers for their helpful comments. This work was partially funded by National Natural Science Foundation of China (No.62476061,62206057,61976056), Shanghai Rising-Star Program (23QA1400200), and Natural Science Foundation of Shanghai (23ZR1403500). The authors would like to thank Huawei Ascend Cloud Ecological Development Project for the support of Ascend 910 processors.

\bibliography{custom}

\appendix
\section{Additional Dataset Details}

\subsection{Chart Types and Topics} \label{app:type_topic}
We predefined several chart types and topics for Self-Instruct prompting. Table~\ref{tab:category} shows the 9 major categories we established, with their corresponding subcategories. 
Additionally, Table~\ref{tab:topics} lists the 38 topics we specified.
It is important to note that these topics do not reflect the actual topic distributions in the generated charts, as we encourage the model to combine and expand upon them.
Regarding the distribution of chart types, we provide a breakdown in Table~\ref{tab:distribution}. 
While we aimed for a roughly balanced representation across different chart types during data construction, some degree of imbalance remains as certain types were more prone to generation errors.

\begin{table*}[h]{
    \centering
    \resizebox{\textwidth}{!}{%
    \begin{tabular}{ll}
        \toprule[1.5pt]
        \textbf{Major Category} & \textbf{Minor Category} \\
        \midrule
        \raisebox{-0.5ex}{\includegraphics[height=0.4cm]{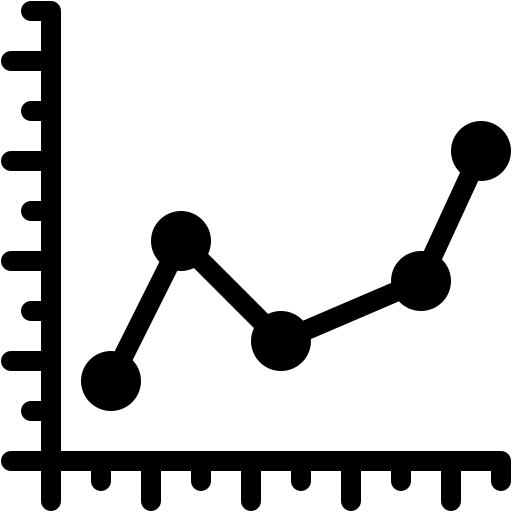}} Line Charts & line chart, line chart with data annotation, line chart with error bar \\
         \raisebox{-0.5ex}{\includegraphics[height=0.4cm]{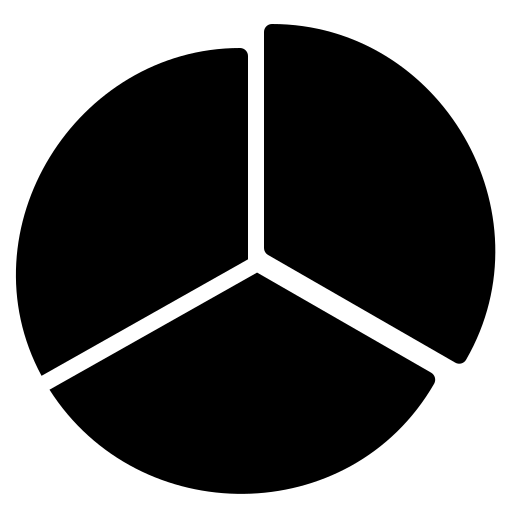}} Pie Charts & pie chart, donut pie chart, sector pie chart, ring chart \\
        \raisebox{-0.5ex}{\includegraphics[height=0.4cm]{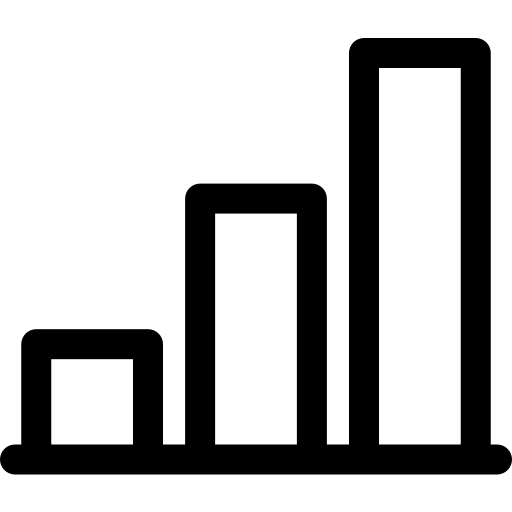}} Bar Charts & bar chart, bar chart with data annotation, stacked bar chart, percentage bar chart, horizontal bar chart \\
        \raisebox{-0.5ex}{\includegraphics[height=0.4cm]{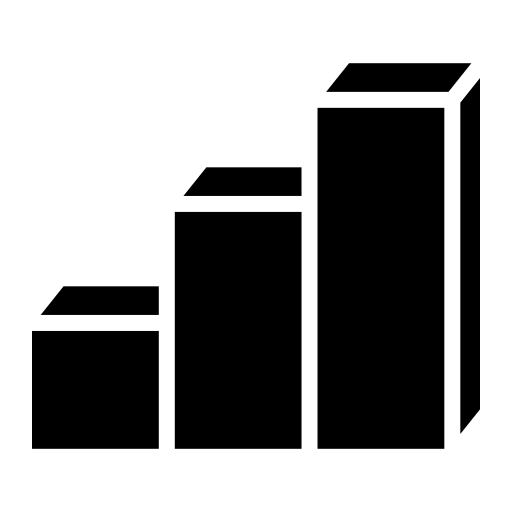}} 3D Bar Charts & 3D bar chart, stacked 3D bar chart, percentage 3D bar chart \\
        \raisebox{-0.5ex}{\includegraphics[height=0.4cm]{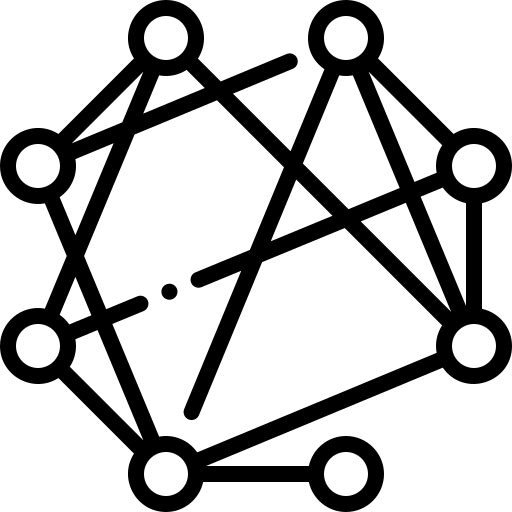}} Node Charts & directed node chart, undirected node chart \\
         \raisebox{-0.5ex}{\includegraphics[height=0.4cm]{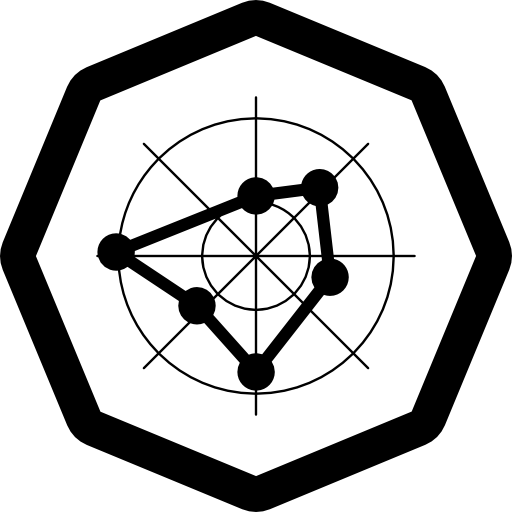}} Radar Charts & radar chart, radar chart with area filling \\
        \raisebox{-0.5ex}{\includegraphics[height=0.4cm]{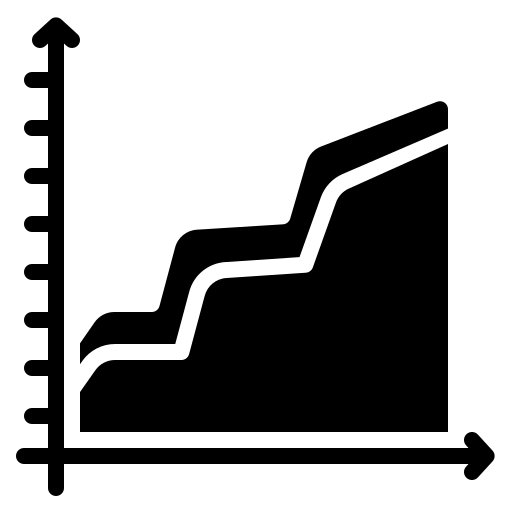}} Area Charts & area chart, stacked area chart \\
        \raisebox{-0.5ex}{\includegraphics[height=0.4cm]{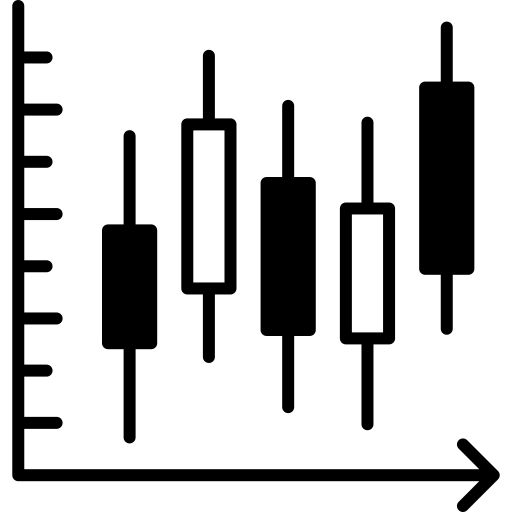}} Box Charts & vertical box chart, horizontal box chart \\
        \raisebox{-0.5ex}{\includegraphics[height=0.4cm]{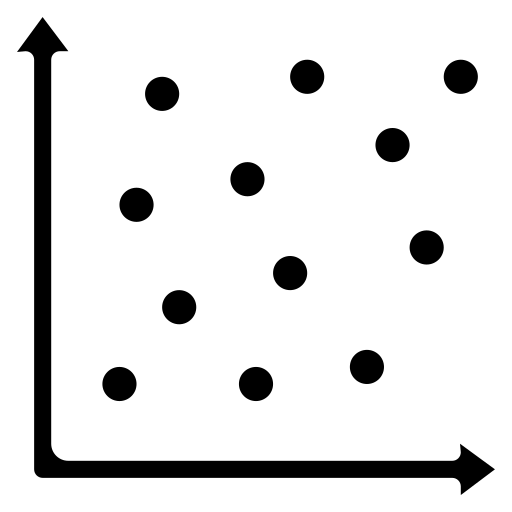}} Scatter Charts & scatter chart, scatter chart with smooth fitting, 3D scatter chart (bubble chart) \\
        \raisebox{-0.5ex}{\includegraphics[height=0.4cm]{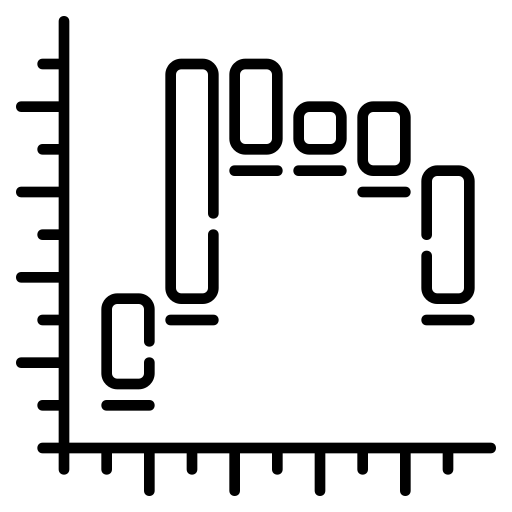}} Specific Charts & heat map, rose chart, funnel chart, waterfall chart, histogram, tree map \\
        \bottomrule[1.5pt]
    \end{tabular}}}
    \caption{Major categories and minor categories of charts in \data.}
    \label{tab:category}
\end{table*}

\begin{table*}[h]
    \centering
    \resizebox{\textwidth}{!}{%
\begin{tabular}{lll}
    \toprule[1.5pt]
    Art and Design & Futurism and Innovation & Agriculture and Food Production \\
    Music and Performance & Astronomy and Space & Transportation and Logistics  \\
    Business and Finance & Social Media and the Web & Real Estate and Housing Market \\
    Travel and Exploration & Society and Community & Government and Public Policy \\
    Books and Publishing & Physics and Chemistry & Education and Academics \\
    Literature and Writing & Energy and Utilities & Environment and Sustainability \\
    History and Culture & Biology and Life Sciences & Language and Communication \\
    Architecture and Building & Retail and E-commerce & Social Sciences and Humanities \\
    Fashion and Style & Religion and Spirituality & Manufacturing and Production \\
    Marketing and Advertising & Food and Beverage Industry & Artificial Intelligence and Robotics \\
    Law and Legal Affairs & Healthcare and Health & Human Resources and Employee Management  \\
    Film and Cinema &  Sports and Entertainment & Computer Science and Information Technology  \\
    Mathematics and Statistics & Science and Engineering & \\
    \bottomrule[1.5pt]
\end{tabular}}
\caption{Predefined chart topics in Self-Instruct prompting.}
\vspace{-10pt}
\label{tab:topics}
\end{table*}

\begin{table*}[h]
\centering
\small
\begin{tabular}{l|ccccccccccc}
\toprule[1.5pt]
\textbf{Types} & Line & Pie & Bar & 3D Bar & Node & Radar & Area & Box & Scatter & Specific & Total \\
\midrule
Train Set & 522 & 415 & 478 & 144 & 120 & 238 & 513 & 331 & 244 & 244 & 3,249 \\
Test Set & 101 & 40 & 112 & 15 & 18 & 19 & 24 & 44 & 63 & 64 & 500 \\
\bottomrule[1.5pt]
\end{tabular}
\caption{Distribution of chart types in \data dataset.}
\label{tab:distribution}
\end{table*}

\subsection{Cost of \data Training Data Construction} \label{app:cost}
Table \ref{tab:cost} provides a detailed expense breakdown. We executed Self-Instruct and Evol-Instruct 3,000 times each to synthesize chart-plotting code, theoretically generating 6,000 charts. However, after accounting for non-executable code and images filtered out by MLLM rating, we ultimately produced 3,249 charts for Q\&A synthesis.

\begin{table*}[h]
\centering
\resizebox{0.85\textwidth}{!}{%
\begin{tabular}{lllll}
\toprule[1.5pt]
\textbf{Step }         & \textbf{Avg. \#tokens of Input}     &\textbf{ Avg. \#tokens of Output} & \textbf{Times} & \textbf{Cost (\$)} \\
\midrule
Self-Instruct & $1,500 + 2,000 = 3,500$    & $500 + 500 = 1,000$     & 3,000 & $\sim 56.25$         \\
Evol-Instruct & $700 + 1,300 = 2,000$      & $300 + 700 = 1,000$     & 3,000 & $\sim 45.00$         \\
Self-Repair   & $500$                      & $500$                   & 1,500 & $\sim 9.38$          \\
Reas-QA-Gen.  & $1,000 + 1,500 \times 4 = 7,000$ & $500 +  300 \times 4 = 1,700$ & 3,249 & $\sim 112.09$        \\
Reco-QA-Gen.  & $800 + 1,200 \times 4 = 5,600$   & $300 + 200 \times 4 = 1,100$  & 3,249 & $\sim 81.23$         \\
\bottomrule[1.5pt]
\end{tabular}}
\caption{The average number of input and output tokens is calculated for each step in the \data construction process. In the equation, each term represents the average number of tokens per step (used only in a multi-step framework), while each multiplier corresponds to the number of times that step is executed. The pricing for GPT-4o-2024-08-06 is \$2.50 per 1M input tokens and \$10.00 per 1M output tokens. As a result, the total cost amounts to approximately \$303.95.}
\label{tab:cost}
\end{table*}

\subsection{Data Contamination Analysis} \label{app:contamination}

\begin{figure*}[h]
    \centering
    \begin{subfigure}{0.45\textwidth}
    \includegraphics[width=\textwidth]{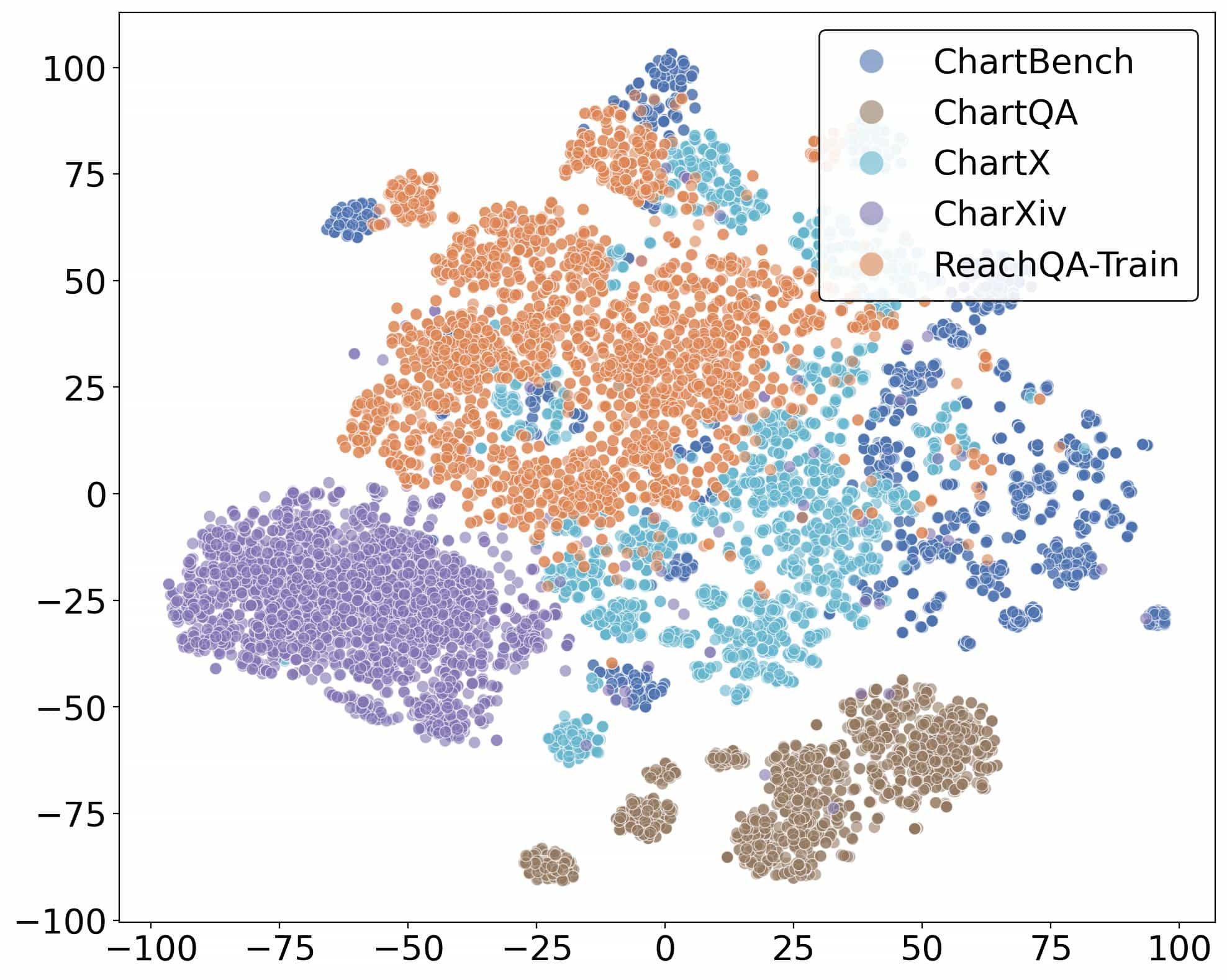}
    \caption{\textbf{Chart: \data vs. Existing datasets}}
    \end{subfigure}
    \hspace{0.05\textwidth}
    \begin{subfigure}{0.45\textwidth}
    \includegraphics[width=\textwidth]{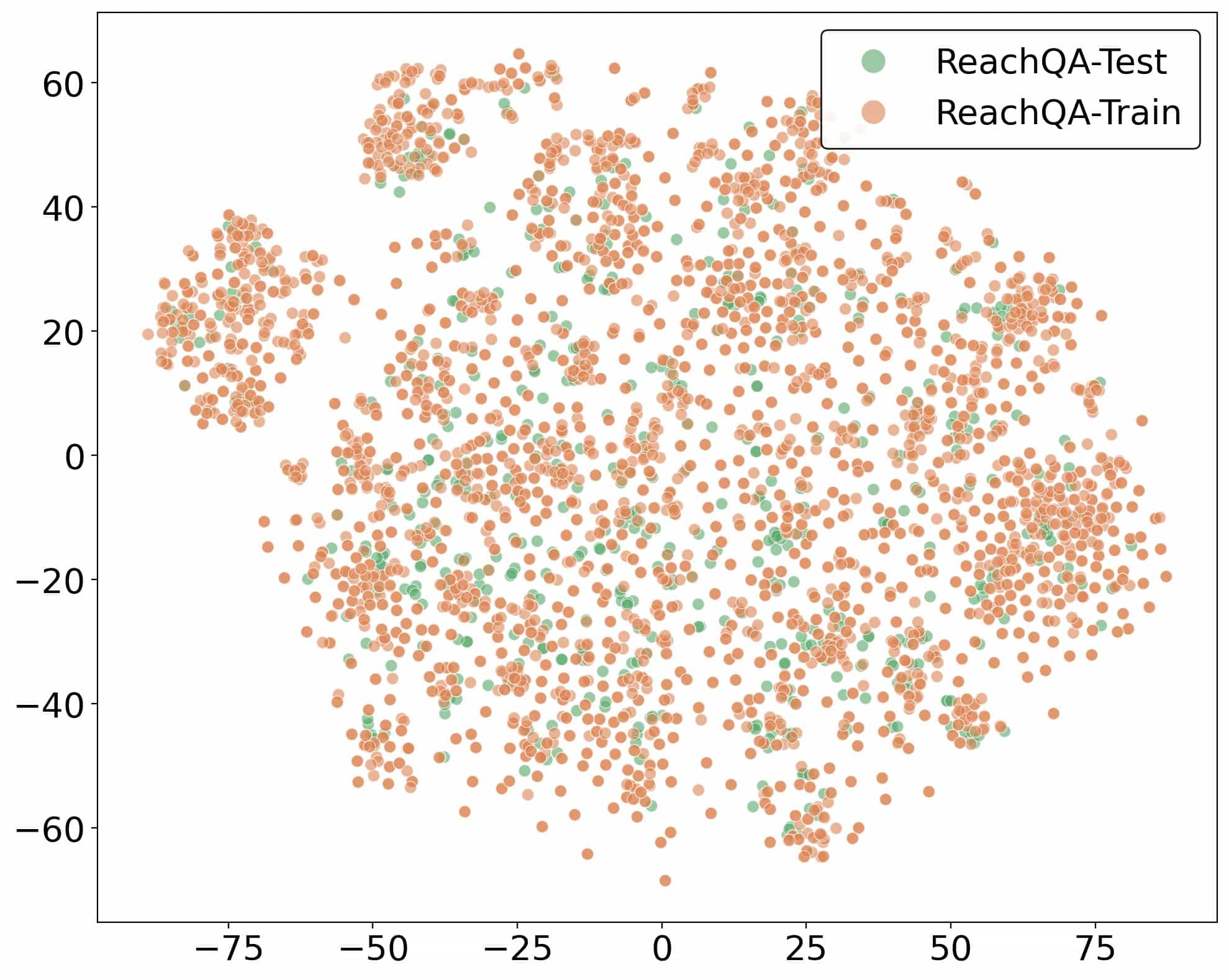}
    \caption{\textbf{Chart: \data Train vs. Test}}
    \end{subfigure}
    \vspace{0.3cm}
    \begin{subfigure}{0.45\textwidth}
    \includegraphics[width=\textwidth]{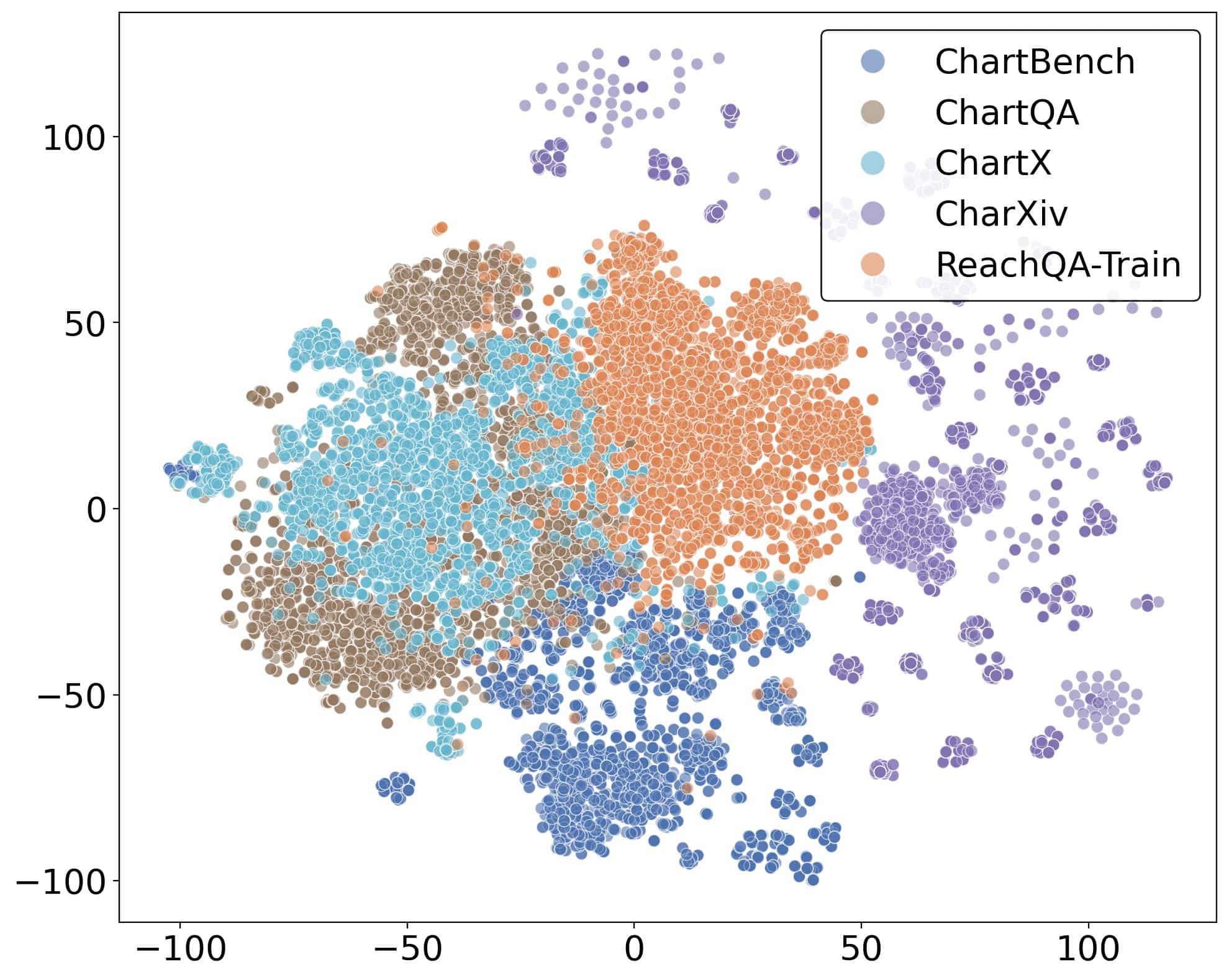}
    \caption{\textbf{Query: \data vs. Existing datasets}}
    \end{subfigure}
    \hspace{0.05\textwidth}
    \begin{subfigure}{0.45\textwidth}
    \includegraphics[width=\textwidth]{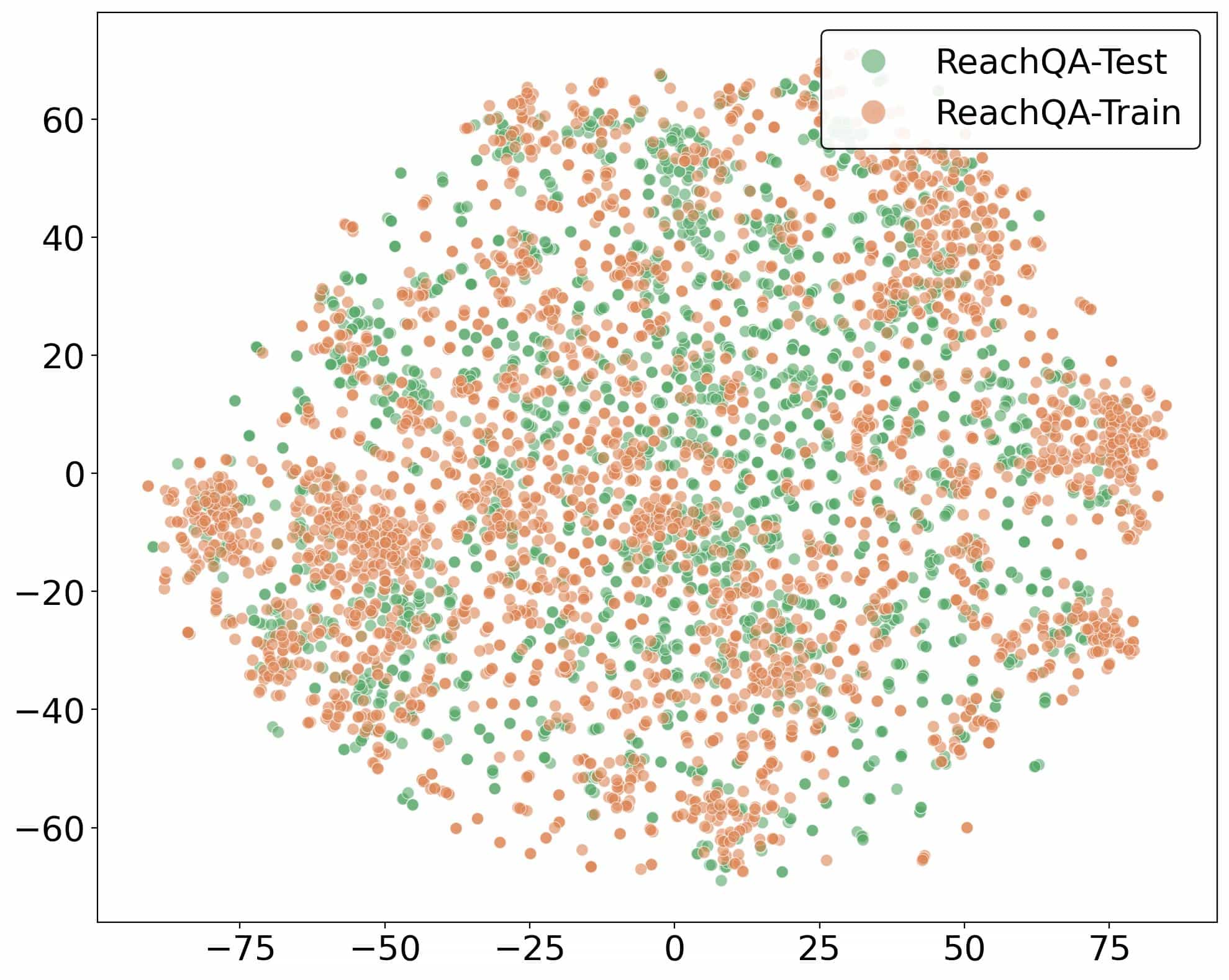}
    \caption{\textbf{Query: \data Train vs. Test}}
    \end{subfigure}
    \caption{\textbf{Data overlap analysis visualization using t-SNE.} We analyze both image-level and query-level similarities through embedding space visualization. (a) and (c) demonstrate the distributional differences between \data and existing datasets, while (b) and (d) examine potential overlap between training and testing splits. The results show clear dataset distinctiveness while revealing expected overlaps due to the shared domain of chart understanding.}
    \label{fig:overlap_analysis}
\end{figure*}

To ensure the validity of our experimental results and exclude potential data contamination, we conduct a comprehensive analysis of data overlap from both dataset-level and split-level perspectives. 
First, to evaluate image-level similarity, we employed the SigLIP-400M encoder~\citep{siglip} to generate embeddings for all chart images across datasets. 
These embeddings were then projected into a two-dimensional space using t-SNE~\citep{tsne} for visualization, following \citet{chartbench}. 
Second, we analyzed query-level similarity using the NV-Embed-v2 model~\citep{nvembed} to generate embeddings for all queries, also visualized through t-SNE.

As shown in Figure~\ref{fig:overlap_analysis}(a) and (c), the visualization demonstrates clear distributional differences between \data and existing chart-related benchmarks. 
While some degree of overlap exists due to the shared nature of chart-related tasks, these instances are limited and do not compromise the overall distinctiveness of our dataset. 
The distinct clustering patterns in both image and query spaces support the validity of our cross-dataset evaluations and confirm that \data presents novel challenges not fully captured by existing benchmarks.

To address potential data leakage between training and testing splits, which were synthesized through the same process, we conduct a more rigorous analysis as visualized in Figure~\ref{fig:overlap_analysis}(b) and (d). 
Beyond visualization, we compute pairwise similarities between all training and testing samples using the chart embeddings. 
Among the identified top $50$ image pairs with similarity scores exceeding $0.9$, our careful manual review revealed only $2$ cases with notable similarities. 
We will exclude them from the test set in future versions and update the evaluation accordingly. 
For the remaining samples, our review confirmed clear differences in chart topics, data values, and query types, ensuring that no further data leakage or contamination is present.

\section{Additional Experiment Details} \label{app:exp}

\subsection{Benchmark Details}

\begin{table*}[ht]
\centering
\resizebox{\textwidth}{!}{%
\begin{tabular}{lcc}
\toprule[1.5pt]
\textbf{Benchmark} & \textbf{Task Focus} & \textbf{Sample Details} \\
\midrule
ChartQA \citep{chartqa}  & Chart Recognition & 2.5k test samples \\
ChartBench \citep{chartbench} & Chart Recognition & 2k binary QA samples and 2.1k numerical QA samples \\
ChartX \citep{chartx}   & Chart Recognition & 6k QA samples \\
\data (ours) & Chart Reco. \& Reas. & 1k recognition-oriented and 1k reasoning-oriented questions \\
CharXiv \citep{charxiv}  & Chart Reco. \& Reas. & 4k descriptive and 1k reasoning questions (validation set) \\
MathVista \citep{MathVista} & General Reasoning & 540 math-targeted and 460 general VQA questions (testmini set) \\
MATH-Vision \citep{mathv} & General Reasoning & 3,040 math competition problems \\
\bottomrule[1.5pt]
\end{tabular}
}
\caption{Summary of benchmarks used in our experiments.}
\label{tab:datasets}
\end{table*}

Table~\ref{tab:datasets} summarizes the benchmarks used in our main experiments, including the number of samples for each dataset. 
Additionally, we use some other popular multimodal datasets in Section~\ref{sec:balancing}, including MME-Reasoning, MME-Perception \citep{mme}, SeedBench \citep{seedbench}, CCBench \citep{mmbench}, POPE \citep{pope}, HallusionBench \citep{hallusionbench}, OCRBench \citep{OCRBench}, and We-Math \citep{We-Math}.

\subsection{Training and Evaluation Details}
For each general open-source model, we conduct supervised fine-tuning (SFT) using our \data training set.
We apply Low-rank Adapters (LoRA, \citealp{lora}) to all linear layers of the language model and projector, with a LoRA rank of 16, a LoRA alpha of 8 and a learning rate of 2e-5.
To fully leverage their capabilities, we prompt all models with a zero-shot CoT prompt, ``\textit{Let's think step by step}'' \citep{zerocot}, following \citet{gpt4o} and \citet{claude3_5}.
Thus, to extract answers from the model responses and assess their correctness, we employ the LLM-as-a-judge method \citep{llm_as_a_judge} to calculate a relaxed accuracy.
The judge model used is GPT-4o, and the prompt template for evaluation can be found in Appendix \ref{prompt:evaluation}.

\section{Visualization of Charts from Different Dataset} \label{app:train_data}
\begin{figure*}[h]
    \centering
    \begin{subfigure}{0.38\textwidth}
        \includegraphics[width=\textwidth]{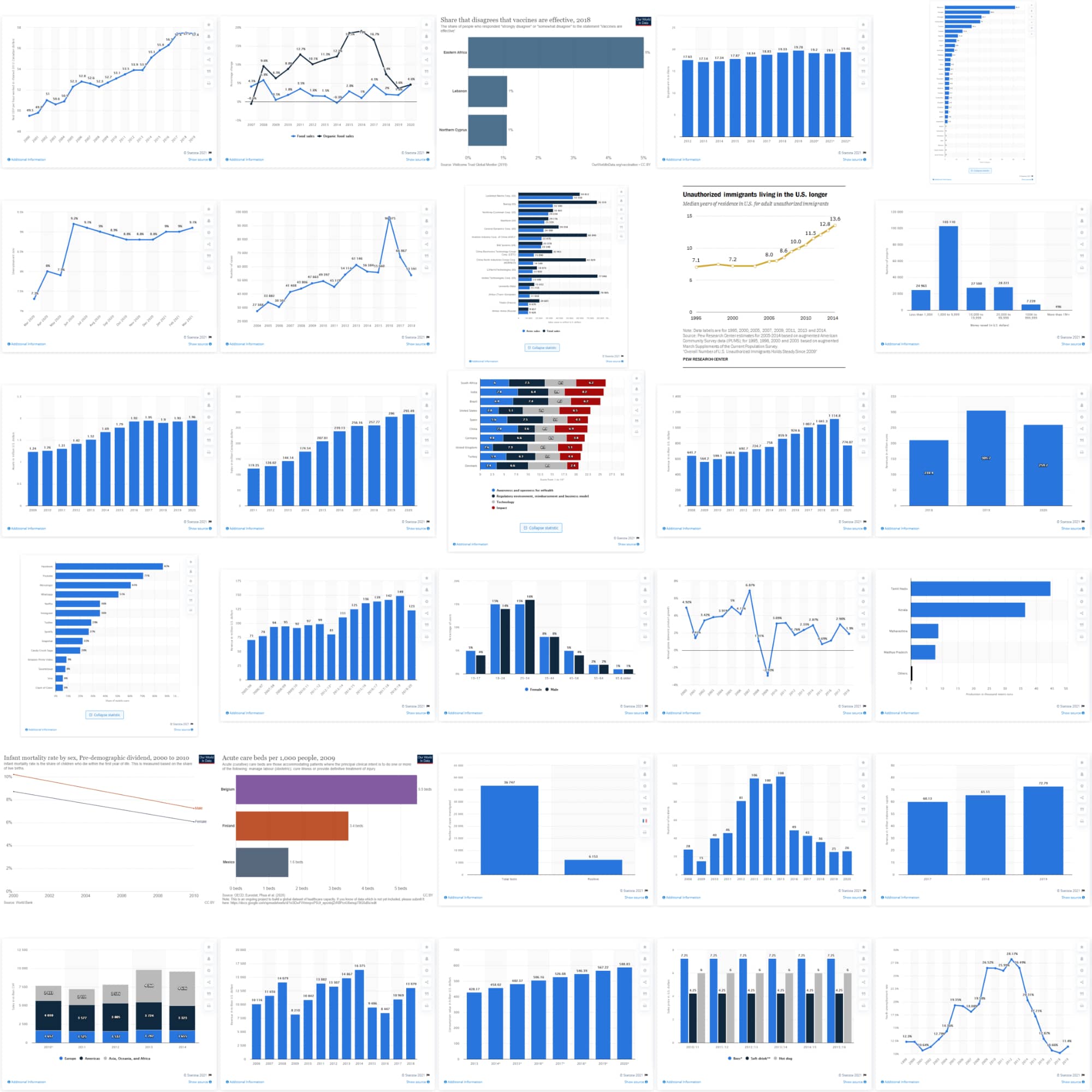}
        \caption{\textbf{ChartQA} contains 3 types of charts collected from 4 websites.}
    \end{subfigure}
    \hspace{0.05\textwidth}
    \begin{subfigure}{0.38\textwidth}
        \includegraphics[width=\textwidth]{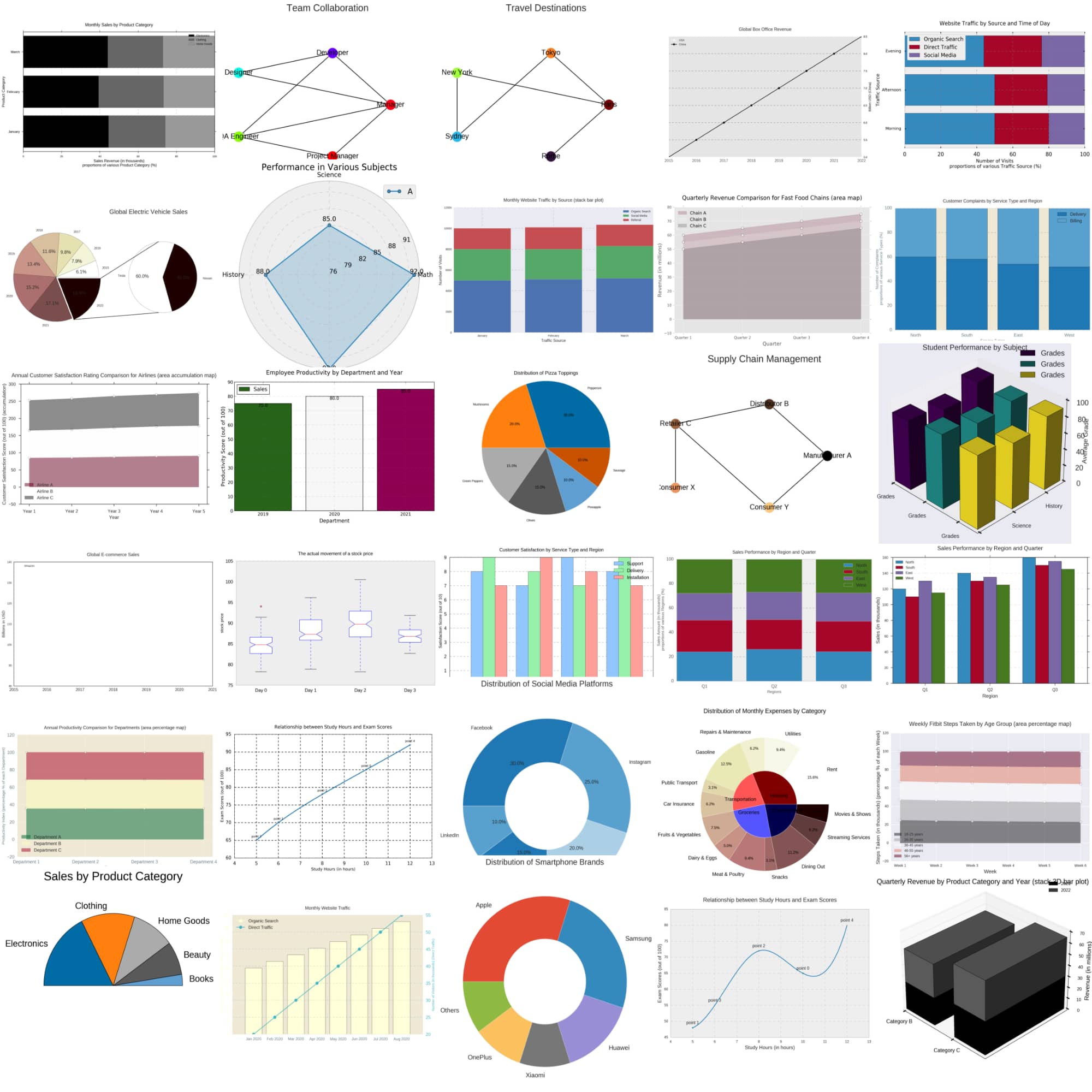}
        \caption{\textbf{ChartBench} contains 9 types of synthetic charts but no visual complexity.}
    \end{subfigure}
    \begin{subfigure}{0.38\textwidth}
        \includegraphics[width=\textwidth]{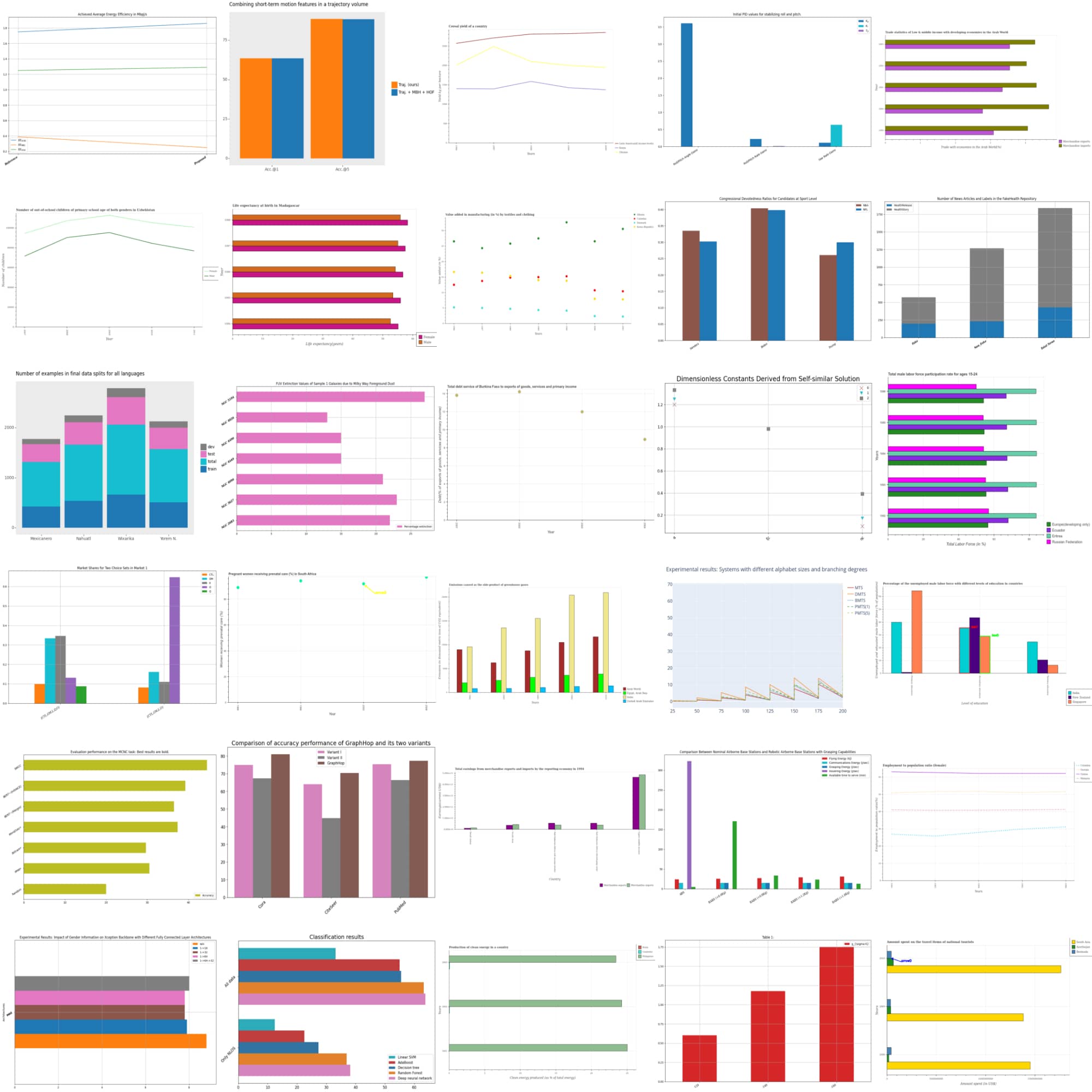}
        \caption{\textbf{ChartAssistant} contains 9 types of synthetic charts but no visual complexity.}
    \end{subfigure}
    \hspace{0.05\textwidth}
    \begin{subfigure}{0.38\textwidth}
        \includegraphics[width=\textwidth]{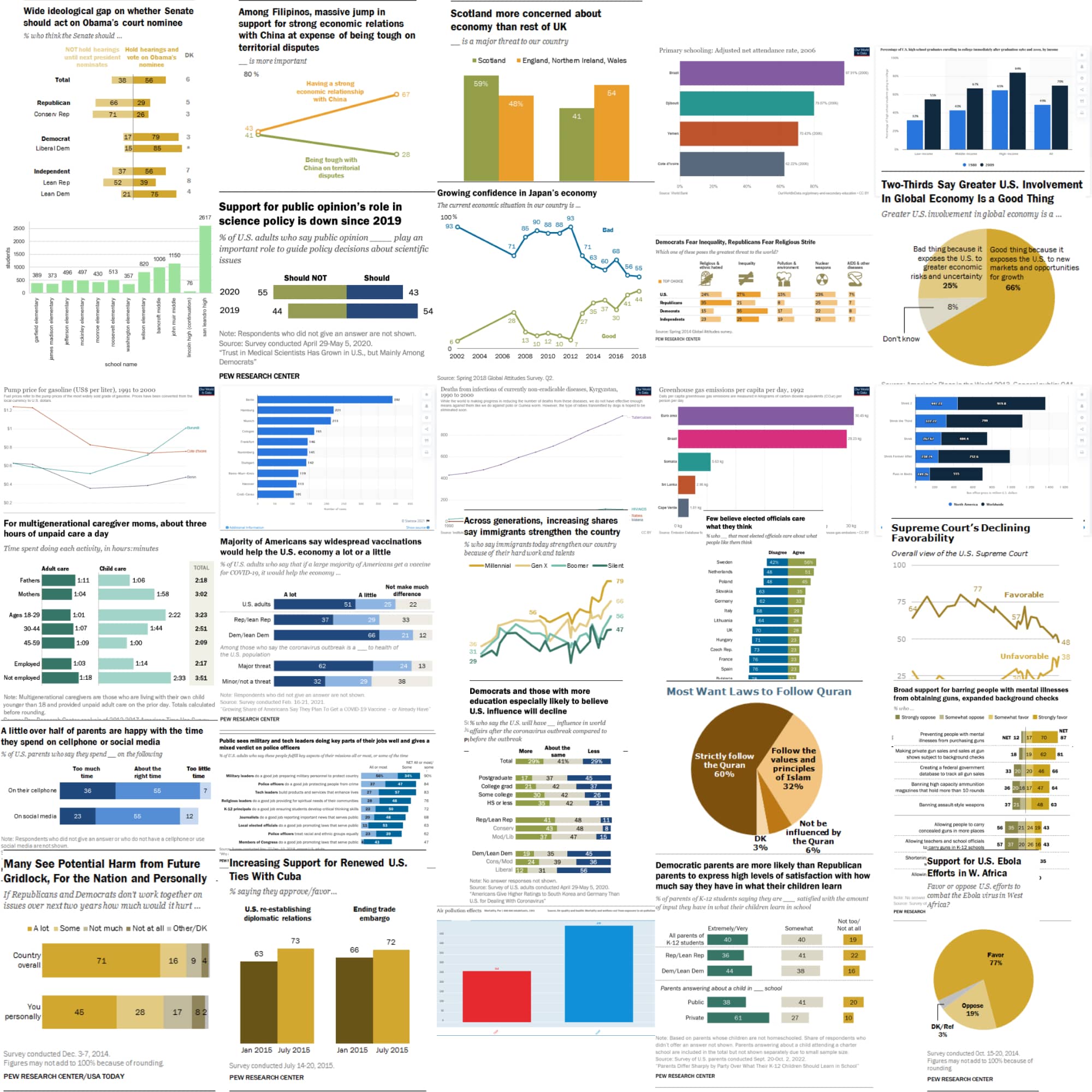}
        \caption{\textbf{ChartGemma} contains charts collected from boarder websites.}
    \end{subfigure}
    \vspace{0.5cm} 
    \begin{subfigure}{0.85\textwidth}
        \includegraphics[width=\textwidth]{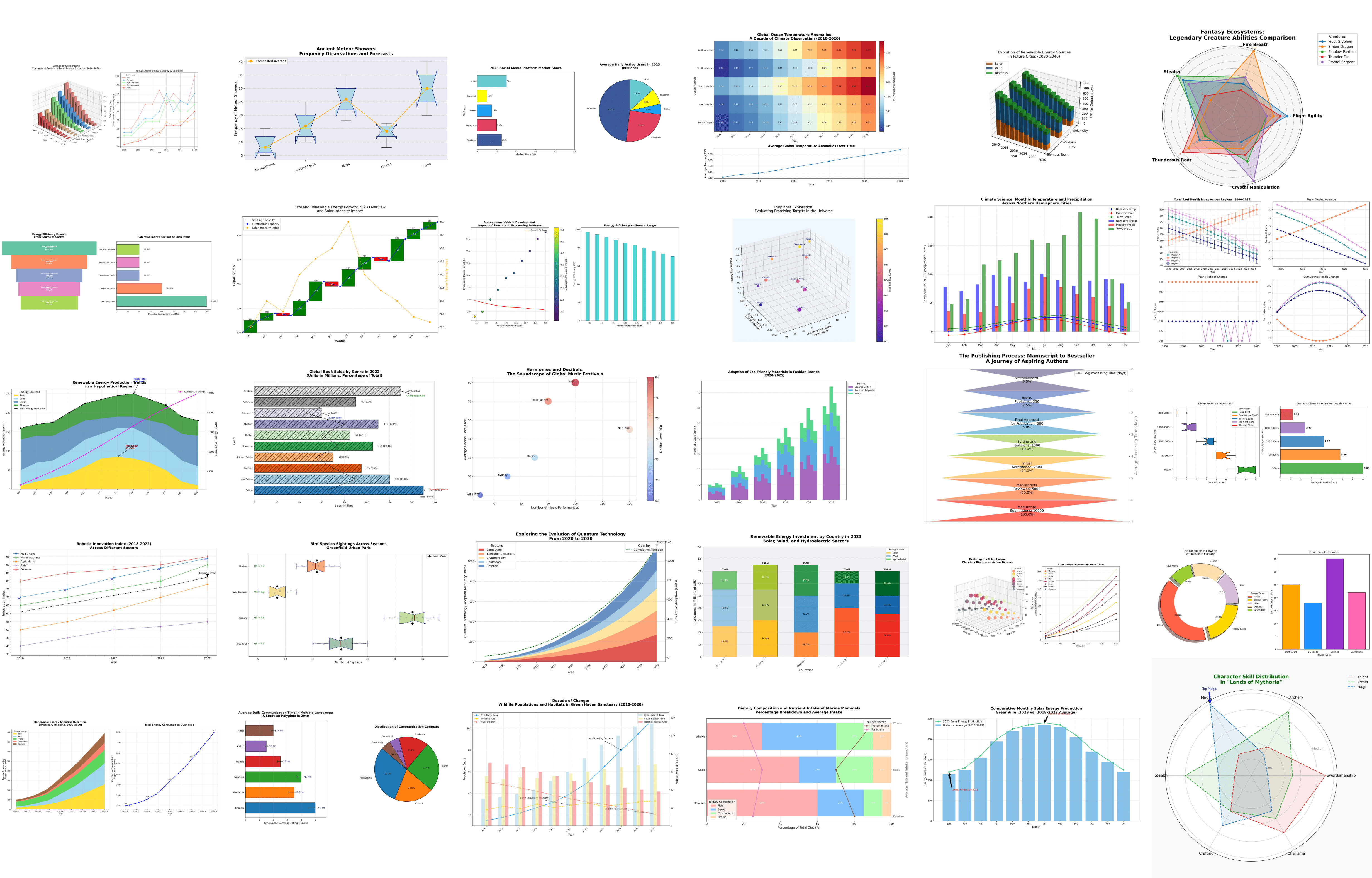}
        \caption{\textbf{\data} contains 10 types of charts and more complex variations.}
    \end{subfigure}
    \vspace{-0.5cm} 
    \caption{\textbf{Visualizations} of different chart-related training datasets. As shown, \data and ChartGemma exhibit higher chart richness compared to several other datasets. But the charts in ChartGemma require manual collection from multiple sources \citep{chartgemma}.}
    \label{fig:vis_training}
\end{figure*}

We randomly sample several charts from the training set of ChartQA \citep{chartqa}, ChartBench \citep{chartbench}, ChartAst \citep{chartast}, ChartGemma \citep{chartgemma}, and \data.
The visualization of the results is presented in Figure~\ref{fig:vis_training}.

\section{Prompt Templates} \label{app:prompt}
We present the prompt templates used in our work.

\subsection{Intermediary Code Synthesis}\label{prompt:code_generation}
The prompts used for code generation via the Self-Instruct method are presented in Figure \ref{prompt:self_instruct}, and Figure \ref{prompt:evol_instruct} shows the prompts for the Evol-Instruct method. As illustrated in Figure \ref{fig:evol_diretion}, we utilize four predefined directions to evolve the simple chart-plotting code.

\begin{figure*}[!ht]
    \centering
\begin{tcolorbox}[colback=gray!5!white,colframe=gray!75!black]
\textbf{User:}
\\~\\
As a MatplotLib expert, you are asked to write a new Python plotting script. This script will be used to generate a type-specific chart with artificial data. Here are the requirements: \\
1. There are several script examples from which you can draw inspiration, but try not to repeat patterns already shown in the examples to maximize diversity. \\
2. Use the Matplotlib library in Python for plotting. You can use auxiliary libraries such as Numpy, but make sure the code works! \\
3. The type of chart you need to plot is \texttt{\{type\}}. Therefore, everything you create must be adapted to fit this type of chart. \\
4. The topic of the chart can be anything you like, for example, \ \texttt{\{topic1\}}, \texttt{\{topic2\}}, etc. \\
5. Based on the given chart type and the topic you choose, you need to construct a suitable backstory, which should be reflected in the title, labels, legend, etc. \\
6. Based on the backstory, construct contextual data inputs in the form of Python lists or Numpy arrays. Information contained in the data can be adapted as appropriate to fit the type of chart. \\
7. You must not use random() to construct the data, as it needs to be explicitly created regardless of your chart type and topic. \\
8. Be as imaginative and creative as possible in drawing the chart, both in terms of data and plotting details. \\
\\
Here are three examples to consider: \\
\texttt{\{demo1\}} \\
\texttt{\{demo2\}} \\
\texttt{\{demo3\}} \\
\\
Now, let's take this task step by step. First, we have to plan out the title and backstory of the chart and create data based on the above.
\\~\\
\textbf{Assistant:} \texttt{\{model\_response\}}
\\~\\
\textbf{User:}
\\~\\
Please complete the entire script by plotting a chart based on the data generated. Here are some highlighted requirements and notes. \\

Requirements: \\
1. If you find that the generated data is not appropriate while plotting the chart, modify it further as needed. \\
2. The information on the chart should be complete enough to be understandable, but avoid including the full backstory or too much text in the figure. \\
3. Avoid occlusion of visual elements. If necessary, automatically adjust the image layout before plt.show() using tight\_layout(). \\
4. If the text in the chart is too long, find a way to make it all visible instead of overlapping. If the title is too long, you can break it into multiple lines. \\
5. Once again, be as imaginative and creative as possible in creating the details of the chart. \\
6. Above all, double-check to ensure the code works. Reduce unnecessary comments and focus on functionality. \\

Now, generate your final plotting script in a single python code block.
\end{tcolorbox}
    \caption{Prompt template for code generation via Self-Instruct method.}
\label{prompt:self_instruct}
\end{figure*}

\begin{figure*}[!ht]
    \centering
\begin{tcolorbox}[colback=gray!5!white,colframe=gray!75!black]
\textbf{User:}
\\~\\
As a MatplotLib expert, you are asked to optimize a Python plotting script to make the plotted chart more complex. The script will be used to generate charts for a mathematical test, so you should make it a little more challenging. \\
\\
This is the code you need to optimize: \\
\texttt{\{code\}} \\

Here's what I'd like you to do to optimize the chart: \texttt{\{direction\}} \\

Now, let's take this task step by step. First, please read the given code carefully and analyze the chart it draws. Then, think about your optimization ideas with the given directions. \\
In this step, you don't need to give the final code, only show the design ideas.
\\~\\
\textbf{Assistant:} \texttt{\{model\_response\}}
\\~\\
\textbf{User:}
\\~\\
Please implement the final optimized script based on the above design ideas combined with the original code. \\

Remember: \\
1. Avoid visual elements that obscure each other, e.g., legends, labels. Automatically adjust the image layout before plt.show() using tight\_layout(). if necessary. \\
2. If the text in the chart is too long, find a way to make all the text show up instead of overlapping. If the title is too long, you can break it into multiple lines. \\
3. Be as imaginative and creative as possible in creating details of the chart, but don't make the chart redundant just to cope. \\
4. If you are adding a new plot, take care that the chart is complete with all the elements, such as labels, axes, legends, and colors, unless it is intended to be shared with the original chart. \\
5. If you are adding a new plot, carefully construct meaningful data and consider whether to give the new sub-plot a sub-title. \\
6. You must not use random() to construct the data, as it needs to be explicitly constructed regardless of your chart type and topic. \\
7. Above all, double-check to make sure the code works. Reduce unnecessary comments and focus on functionality. \\

Now, generate your optimized plotting script in a single python code block.
\end{tcolorbox}
    \caption{Prompt template for code generation via Evol-Instruct method.}
    \label{prompt:evol_instruct}
\end{figure*}

\begin{figure*}[!ht]
    \centering
\begin{tcolorbox}[colback=gray!5!white,colframe=gray!75!black]
\textbf{Evolution Direction:}
\begin{itemize}[itemsep=0pt, left=0pt]
    \item Increase the size of the input data or the number of data groups as appropriate so that it requires a higher level of mathematical understanding. Note if there is a sum requirement.
    \item Try changing or adding some visual elements to make visual effect better. The elements you add must make sense and not be redundant.
    \item Incorporate an overlay plot of a different type on the original chart. Use related but not identical data for the added plot.
    \item Extend an additional subplot of a different type beside the original chart (2 in total). Use related but not identical data for the added plot.
\end{itemize}
\end{tcolorbox}
    \caption{Predefined evolution directions for Evol-Instruct method.}
    \label{fig:evol_diretion}
\end{figure*}

\subsection{Bi-directional Translation}\label{prompt:instruction_generation}

The prompt used for the Self-Repair method is presented in Figure \ref{prompt:self_repair}. 
Additionally, the prompt templates for generating reasoning-oriented questions and answers are listed in Figure \ref{prompt:reason_question} and Figure \ref{prompt:reason_answer}. 
The prompt details for generating recognition-oriented questions and answers are listed in Figure \ref{prompt:recognition_question} and Figure \ref{prompt:recognition_answer}.

\begin{figure*}[!ht]
    \centering
\begin{tcolorbox}[colback=gray!5!white,colframe=gray!75!black]
\textbf{User:}
\\~\\
As a Python and Matplotlib expert, you have been asked to fix the following code. \\ \\
The error code is: \\
\texttt{\{code\}}  \\
\\
The code reports the following error message when run: 
\texttt{\{error\}} \\

Please analyze the error first, and then provide the revised code within a single Python code block. There should only be one Python code block in your response, containing the complete revised code.
\end{tcolorbox}
    \caption{Prompt template for Self-Repair.}
    \label{prompt:self_repair}
\end{figure*}

\begin{figure*}[!ht]
    \centering
\begin{tcolorbox}[colback=gray!5!white,colframe=gray!75!black]
\textbf{User:}
\\~\\
You are both an expert Matplotlib plotter and a professional maths teacher. Now, you are asked to generate a mathematical reasoning question about a given chart. This chart and question will be used as a question on this year's college admissions examination. As a question writer, you need to ensure that the question is challenging yet fair, testing the student's ability to analyze data, interpret trends, and apply mathematical concepts. \\ \\
First, please read the following plotting script in Python, try to visualize the figure in your mind and understand the meaning of the chart. After you've analyzed this chart, we'll start generating the associated question. \\ \\
Here are some tips for you: \\ 
1. The plotting script (including the code itself, data mapping and labels) is absolutely correct, and you can trust it completely. \\
2. The question needs to be based on the chart type, chart topic, and the given data. It can relate to the chart as a whole or to localized details, so you need to look closely. \\
3. The question should be challenging, requiring visual observation skills and mathematical reasoning skills. So, you need to have a deep understanding of the chart. \\
4. If there is no data annotation in the figure, try not to generate questions that require too much numerical recognition to reduce inconsistent answers due to visual errors. \\
5. If some numerical recognition is needed, choose distinguishable colors, lines, heights, and other features that make it easy to estimate without data annotation. \\
6. You don't need to describe what the chart shows in the question text, including values, labels, etc. This can be left to the student to recognize. \\ \\
Here is the plotting script: \\
\texttt{\{code\}} \\ \\
Now, please generate 4 questions at a time, each of which needs to look at a different aspect of the chart. \\
Your output needs to follow this JSON format, and no other text included:\\
\{``question\_list'': [``the question you generate'']\}
\end{tcolorbox}
    \caption{Prompt template for generating reasoning-oriented questions.}
    \label{prompt:reason_question}
\end{figure*}

\begin{figure*}[!ht]
    \centering
\begin{tcolorbox}[colback=gray!5!white,colframe=gray!75!black]
\textbf{User:}
\\~\\
You are both a Matplotlib graphing expert and a professional math teacher. Now, you have been asked to generate an answer to a given chart and question. This chart and question will be used as a question on this year's college admissions examination. As the answer writer, you need to ensure that the answer is correct, detailed, and educational. \\ \\
First, please read the following plotting script in Python, try to visualize the figure in your mind and understand the meaning of the chart. After you've analyzed this chart, we'll start generating the answer. \\ 

Here is the plotting script: \\
\texttt{\{code\}} \\

Here are some tips for you to generate the answer: \\
1. First and foremost, the answer needs to be based on the chart information. \\
2. In the answer, you will also need to solve the question step-by-step, including reasoning steps and recognition steps (but keep concise). \\
3. You need to explicitly involve a final answer; the type of answer can be a certain number, a noun, or Yes/No, etc. \\
4. The answer should contain multiple reasoning or calculation steps and be presented in an understandable and educational paragraph. \\
5. NEVER include any information relating to the Python script in the answer text, as students will ONLY have access to the plotted figure. \\ 

Here is the question: \texttt{\{question\}} \\ \\ 
Your output needs to follow this JSON format, and no other text should be included: \\
\{``analysis'': ``your analysis about the scirpt and question'', ``answer'': ``your step-by-step answer''\}
\end{tcolorbox}
    \caption{Prompt template for generating reasoning-oriented answers.}
    \label{prompt:reason_answer}
\end{figure*}

\begin{figure*}[!ht]
    \centering
\begin{tcolorbox}[colback=gray!5!white,colframe=gray!75!black]
\textbf{User:}
\\~\\
You are both an expert Matplotlib plotter and a professional maths teacher. Now, you are asked to generate a recognition-oriented question about a given chart. This chart and question will be used as a question on this year's elementary math examination to test students' ability to read charts. \\ \\
First, please read the following plotting script in Python, try to visualize the figure in your mind and understand the meaning of the chart. After you've analyzed this chart, we'll start generating the associated question. \\ \\
Here are some tips for you: \\
1. The plotting script (including the code itself, data mapping, and labels) is absolutely correct and you can trust it completely. \\
2. Descriptive questions are questions that can be answered based on basic chart information, such as titles, labels, tick marks, colors, etc. \\
3. The generated Q\&A needs to be based on the chart type and data. It should be answerable through visual observation. \\
4. If there is no data annotation in the figure, try not to generate questions that require too many numerical recognitions to reduce inconsistent answers due to visual errors. \\
5. If some numerical recognition is needed, choose distinguishable colors, lines, heights, and other features that make it easy to estimate without data annotation. \\
6. You don't need to describe the content of the figure in the question text. This can be left for students to think about. \\
7. This question needs to explicitly involve a final answer; the type of answer can be a certain number, a noun, or Yes/No, etc. \\
8. NEVER include any information relating to the Python script in the question or answer, as students will ONLY have access to the plotted figure. \\ \\
Here are some examples of recognition-oriented questions: \\
- How many colors are used in the chart? How many city categories are in the chart? \\
- What's the leftmost value of the bar in China? And what is the value of the bar next to it? \\
- For the subplot at row 2 and column 1, what is the minimum value of the solid line? \\
- Which name does the second-largest sector represent? What is its value? \\
- Does the blue triangle in the chart represent a higher value than the red circle? \\ \\
Here is the plotting script: \\
\texttt{\{code\}} \\ \\
Now, please generate 4 questions at a time, each of which needs to look at a different aspect of the chart. \\
Your output needs to follow this JSON format, and no other text included:\\
\{``question\_list'': [``the question you generate'']\}
\end{tcolorbox}
    \caption{Prompt template for generating recognition-oriented questions.}
    \label{prompt:recognition_question}
\end{figure*}

\begin{figure*}[!ht]
    \centering
\begin{tcolorbox}[colback=gray!5!white,colframe=gray!75!black]
\textbf{User:}
\\~\\
You are both a Matplotlib graphing expert and a professional math teacher. Now, you have been asked to generate an answer to a given chart and question. This chart and question will be used as a question on this year's elementary math examination to test students' ability to read charts. As the answer writer, you need to ensure that the answer is correct, detailed, and educational. \\

First, please read the following plotting script in Python, try to visualize the figure in your mind and understand the meaning of the chart. After you've analyzed this chart, we'll start generating the answer. \\

Here is the plotting script: \\
\texttt{\{code\}} \\ \\
Here are some tips for you to generate the answer: \\
1. First and foremost, the answer needs to be based on the chart information. \\
2. In the answer, you will also need to solve the question step-by-step, including reasoning steps and recognition steps (but keep concise). \\
3. You need to explicitly involve a final answer; the type of answer can be a certain number, a noun, or Yes/No, etc. \\
4. The answer should contain multiple reasoning or calculation steps and be presented in an understandable and educational paragraph. \\
5. NEVER include any information relating to the Python script in the answer text, as students will ONLY have access to the plotted figure. \\ \\
Here is the question: \texttt{\{question\}} \\ \\
Your output needs to follow this JSON format, and no other text should be included: \\
\{``analysis'': ``your analysis about the scirpt and question'', ``answer'': ``your step-by-step answer''\}
\end{tcolorbox}
    \caption{Prompt template for generating recognition-oriented answers.}
    \label{prompt:recognition_answer}
\end{figure*}

\subsection{Quality Assurance}\label{prompt:rating}
The prompt details for rating charts and Q\&A are illustrated in Figure \ref{prompt:rating_chart} and \ref{prompt:rating_qa}.

\begin{figure*}[!ht]
    \centering
\begin{tcolorbox}[colback=gray!5!white,colframe=gray!75!black]
\textbf{User:}
\\~\\
\texttt{<image>} \\ \\
You are a strict MatplotLib plotter and have been asked to evaluate the given chart. Rate the chart from 1 to 5 based on these criteria: \\
\\
\textbf{1 point}: This chart is the poorest in quality and fails to accurately represent any relevant data. It is characterized by a complete breakdown in visual representation; elements are cluttered, text heavily overlaps, legend is missing, or large areas are left blank, making the chart unreadable. The design shows no understanding of effective data visualization practices.

\textbf{2 points}: The chart displays incorrect or irrelevant visual elements, with significant inaccuracies that misrepresent the data. The layout suffers from clutter, substantial overlapping of text and other visual elements, such as the legend or labels, and poorly designed axes that result in uneven distribution, severely impeding accurate interpretation.

\textbf{3 points}: This chart represents some correct data points but makes basic errors in visual representation. It may use misleading scales, inappropriate chart types, omit key data. Visual clutter and overlapping elements, such as text obscuring parts of the chart or sub-diagrams overlapping each other, detract from the chart's clarity and readability.

\textbf{4 points}: The chart accurately represents most of the major data points and important details of the dataset. Minor visual errors exist, such as slight occlusions of text or sub-optimal positioning of elements like legends or labels, but these do not significantly affect the overall accuracy or readability. The chart demonstrates a good understanding of effective visualization techniques but could still be improved in terms of visual layout and the balance of details.

\textbf{5 points}: This is an exemplary chart that perfectly encapsulates all critical data points and relationships with outstanding visual clarity and no occlusions. It demonstrates a thorough understanding of data visualization techniques, making excellent use of space and visual elements. The chart is informative, clear, engaging, and free from any visual errors. \\ \\
Score the chart on this scale, providing a short analysis and a single value. Your response should be in the format: \\
Analysis: (your analysis) \\
Rating: (int)
\end{tcolorbox}
    \caption{Prompt template for rating the chart quality.}
    \label{prompt:rating_chart}
\end{figure*}

\begin{figure*}[!ht]
    \centering
\begin{tcolorbox}[colback=gray!5!white,colframe=gray!75!black]
\textbf{User:}
\\~\\
\texttt{<image>} \\ \\
You are a visual question answering (VQA) data annotator. Your task is to review the following chart and question, and determine if the answer is correct based on the information in the chart. You should carefully analyze the chart, taking into account all relevant data points, labels, and trends. Then, conduct an in-depth analysis to determine if there are any unreasonable or incorrect aspects in the figure, question, or answer. \\

Specifically, consider the following points: \\
1. Are the provided question and answer relevant to the chart? Can the answer be found in the chart? \\
2. Do the colors in the charts and questions correspond correctly? Are there instances where the colors are incorrectly referred to? \\
3. Do the data in the charts and questions correspond correctly? Are there any errors in the data or misalignment of information? \\
4. Is the provided answer correct? Are there any logical errors or unreasonable points? \\
5. Apart from the points listed above, is there anything else in this question and answer that doesn't make sense? \\

Here is the question and answer about the given chart: \\
Question: \texttt{\{question\}} \\
Answer: \texttt{\{answer\}} \\
\\
You are asked to provide a short analysis and decide whether to keep the example. Your response should be in the format: \\
Analysis: (your analysis) \\
Decision: (yes/no)
\end{tcolorbox}
    \caption{Prompt template for rating Q\&A quality.}
    \label{prompt:rating_qa}
\end{figure*}

\subsection{Evaluation}\label{prompt:evaluation}
In the evaluation process, we utilize the LLM-as-a-judge method. The detailed prompt template is illustrated in Figure \ref{prompt:llm_as_a_judge}.

\begin{figure*}[!ht]
    \centering
\begin{tcolorbox}[colback=gray!5!white,colframe=gray!75!black]
\textbf{User:}
\\~\\
Compare the ground truth with the prediction from AI model and determine if the prediction is correct. The question is about an image, which we have not given here. You need to determine whether the model's prediction is consistent with the ground truth. No points will be awarded for wrong answers, over answers or under answers. The reasoning process in the prediction does not need to be considered too much, you only need to determine if the final answer is consistent. There are times when the answer may have a different form of expression and some variation is acceptable.\\ \\
\#\# Question: \texttt{\{question\}} \\
\#\# Ground Truth: \texttt{\{answer\}} \\
\#\# Prediction: \texttt{\{prediction\}} \\ \\
Now, let's analyze it and then provide your judgment. Your response must follow the format below: \\
Analysis: (analyze the correctness briefly) \\
Correctness: (Yes or No)
\end{tcolorbox}
    \caption{Prompt template for evaluating the model prediction with LLMs.}
    \label{prompt:llm_as_a_judge}
\end{figure*}

\end{document}